\documentclass[10pt,twocolumn,letterpaper]{article}

\usepackage{cvpr}
\usepackage{times}
\usepackage{epsfig}
\usepackage{graphicx}
\usepackage{amsmath}
\usepackage{amssymb}
\usepackage[utf8]{inputenc}
\usepackage[caption=false]{subfig}

\usepackage[pagebackref=true,breaklinks=true,letterpaper=true,colorlinks,bookmarks=false]{hyperref}

\cvprfinalcopy % *** Uncomment this line for the final submission

\ifcvprfinal\fi
\begin{document}

\title{PlaNet - Photo Geolocation with Convolutional Neural Networks}

\author{Tobias Weyand\\
Google\\
{\tt\small weyand@google.com}
\and
Ilya Kostrikov\\
RWTH Aachen University\\
{\tt\small ilya.kostrikov@rwth-aachen.de}
\and
James Philbin\\
Google\\
{\tt\small philbinj@gmail.com}
}

\maketitle

\graphicspath{{images/}}

\newcommand{\PAR}[1]{\vskip4pt \noindent {\bf #1}}

% Basic macros
\makeatletter
\DeclareRobustCommand\onedot{\futurelet\@let@token\@onedot}
\def\@onedot{\ifx\@let@token.\else.\null\fi\xspace}
\def\eg{\emph{e.g}\onedot} \def\Eg{\emph{E.g}\onedot}
\def\ie{\emph{i.e}\onedot} \def\Ie{\emph{I.e}\onedot}
\def\cf{\emph{cf}\onedot} \def\Cf{\emph{Cf}\onedot}
\def\etc{\emph{etc}\onedot} \def\vs{\emph{vs}\onedot}
\def\wrt{w.r.t\onedot} \def\dof{d.o.f\onedot}
\def\etal{\emph{et al}\onedot}
\def\vs{vs\onedot}
\makeatother

\begin{abstract}
Is it possible to build a system to determine the location where a photo was
taken using just its pixels? In general, the problem seems
exceptionally difficult: it is trivial to construct situations where
no location can be inferred. Yet images often contain informative cues
such as landmarks, weather patterns, vegetation, road markings, and
architectural details, which in combination may allow one to determine
an approximate location and occasionally an exact location. Websites
such as \emph{GeoGuessr} and \emph{View from your Window} suggest that humans are
relatively good at integrating these cues to geolocate images,
especially en-masse.
In computer vision, the photo geolocation problem is usually approached using image retrieval methods.
In contrast, we pose the problem as one of
classification by subdividing the surface of the earth into thousands
of multi-scale geographic cells, and train a deep network using
millions of geotagged images. While previous approaches
only recognize landmarks or perform approximate matching using global
image descriptors, our model is able to use and integrate multiple
visible cues. We show that the resulting model, called PlaNet,
outperforms previous approaches and even attains superhuman levels of
accuracy in some cases.
Moreover, we extend our model to photo albums by combining it with a
long short-term memory (LSTM) architecture. By learning to exploit
temporal coherence to geolocate uncertain photos, we demonstrate that
this model achieves a 50\% performance improvement over the
single-image model.
\end{abstract}

\section{Introduction}

\begin{figure}[t]
  \centering
  \setlength\tabcolsep{1pt}
   \def \si {.38}
   \def \sm {.55}
  \def \sc {.01}
  
    \subfloat[]{\shortstack{\includegraphics[width=\si\linewidth]{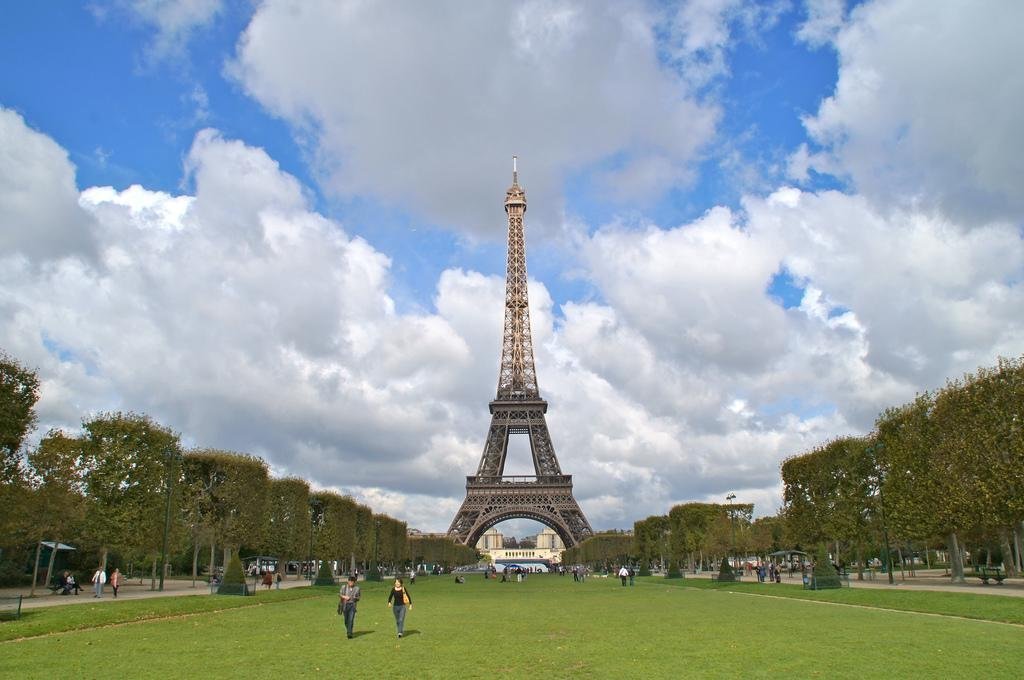} \\ \fontsize{3pt}{1em}\selectfont Photo CC-BY-NC by stevekc} 
    \includegraphics[width=\sm\linewidth]{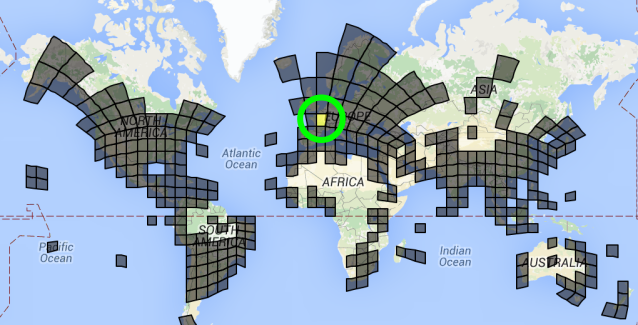} 
    \includegraphics[width=\sc\linewidth]{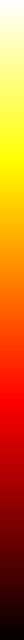}}
    \vspace{-5pt}
    \subfloat[]{\shortstack{\includegraphics[width=\si\linewidth]{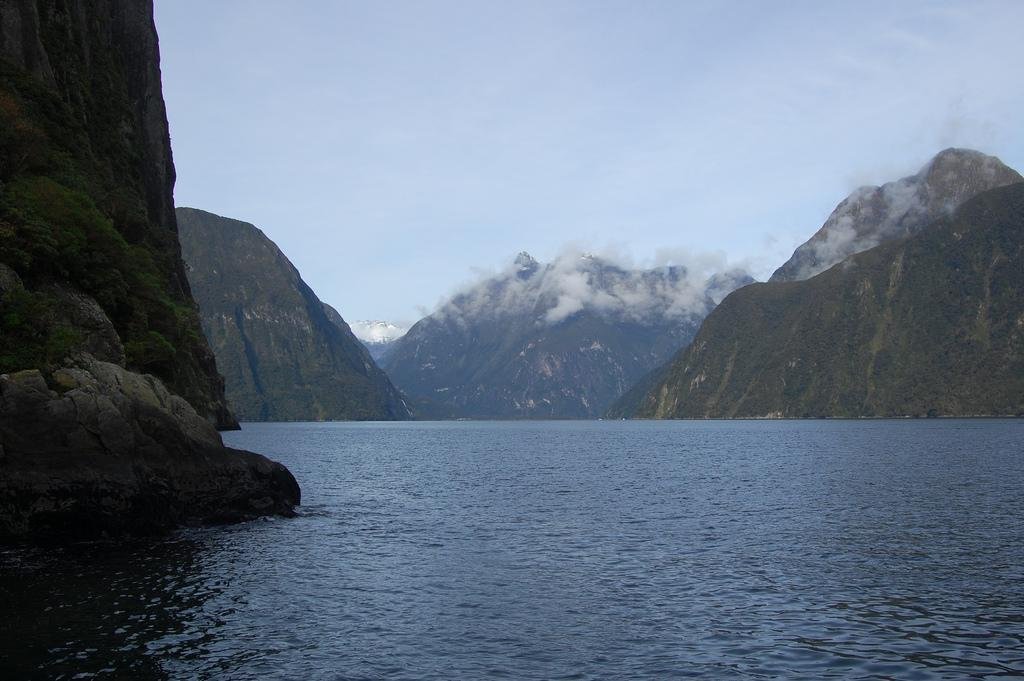} \\ \fontsize{3pt}{1em}\selectfont Photo CC-BY-NC by edwin.11} 
    \includegraphics[width=\sm\linewidth]{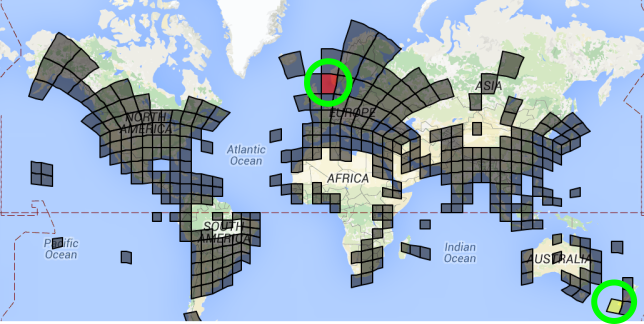} 
    \includegraphics[width=\sc\linewidth]{qualitative/uncertainty/colormap.jpg}} 
   \vspace{-5pt}
       \subfloat[]{\shortstack{\includegraphics[width=\si\linewidth]{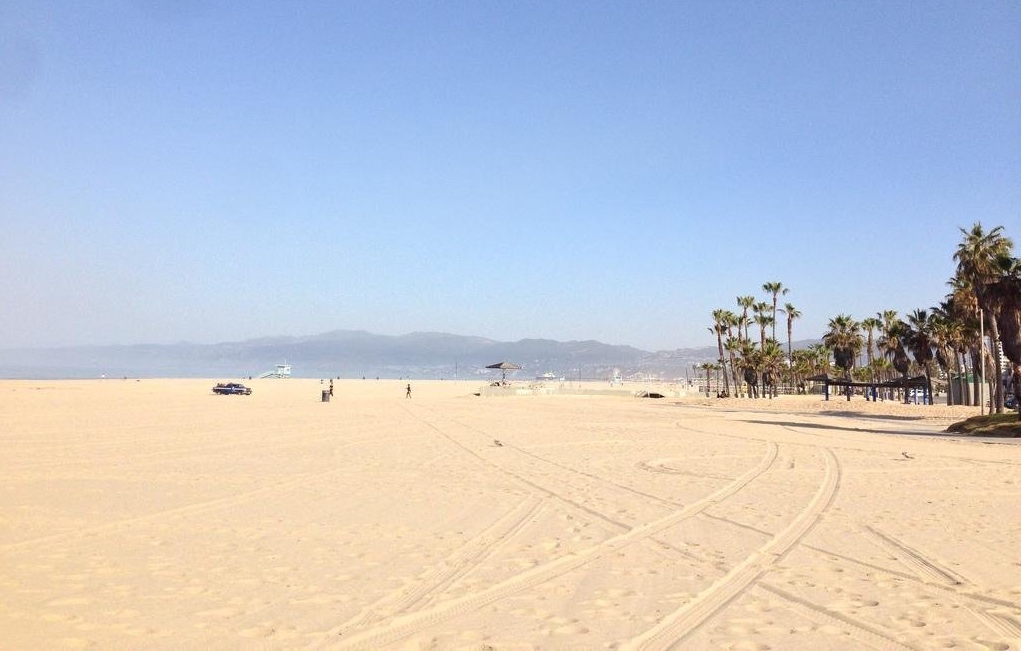} \\ \fontsize{3pt}{1em}\selectfont Photo CC-BY-NC by jonathanfh} 
    \includegraphics[width=\sm\linewidth]{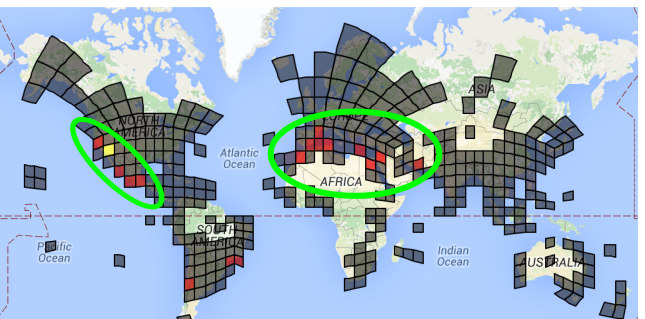} 
    \includegraphics[width=\sc\linewidth]{qualitative/uncertainty/colormap.jpg}}
  \caption{Given a query photo (left), PlaNet outputs a
probability distribution over the surface of the earth (right).
Viewing the task as a classification problem allows PlaNet to
express its uncertainty about a photo. While the Eiffel Tower (a) is
confidently assigned to Paris, the model believes that the fjord photo (b)
could have been taken in either New Zealand or Norway. For the beach
photo (c), PlaNet assigns the highest probability to southern
California (correct), but some probability mass is also assigned to
places with similar beaches, like Mexico and the Mediterranean. (For
visualization purposes we use a model with a much lower spatial
resolution than our full model.)}
  \label{fig:uncertainty}
\end{figure}

Photo geolocation is an extremely challenging task since many photos
offer only a few cues about their location and these cues can often be
ambiguous. For instance, an image of a typical beach scene could be
taken on many coasts across the world. Even when landmarks are present
there can still be ambiguity: a photo of the Rialto Bridge could be
taken either at its original location in Venice, Italy, or in Las
Vegas which has a replica of the bridge! In the absence of obvious and
discriminative landmarks, humans can fall back on their world
knowledge and use multiple cues to infer the location of a photo. For
example, the language of street signs or the driving direction of cars
can help narrow down possible locations. Traditional computer vision
algorithms typically lack this kind of world knowledge, relying on the
features provided to them during training.

Most previous work has therefore focused on
covering restricted subsets of the problem, like landmark
buildings~\cite{Avrithis10MM,Quack08CIVR,Zheng09CVPR}, cities where
street view imagery is
available~\cite{Chen11CVPR,Kim15ICCV,Zamir14PAMI}, or places where
coverage of internet photos is dense enough to allow building a
structure-from-motion reconstruction that a query photo can be matched
against~\cite{Li12ECCV,Sattler12ECCV}.  In contrast, our goal is to
localize any type of photo taken at any location. To our knowledge,
very few other works have addressed this
task~\cite{Hays08CVPR,Hays14MLEVI}.

We treat the task of geolocation as a classification
problem and subdivide the surface of the earth into a set of
geographical cells which make up the target classes. We then train a convolutional neural network
(CNN) \cite{Szegedy15CVPR} using millions of geotagged images. At inference time, our model
outputs a discrete probability distribution over the earth,
assigning each geographical cell a likelihood that the input
photo was taken inside it.

The resulting model, which we call \emph{PlaNet}, is capable of
localizing a large variety of photos. Besides landmark buildings and
street scenes, PlaNet can often predict the location of nature
scenes like mountains, waterfalls or beaches with surprising accuracy.
In cases of ambiguity, it will often output a distribution
with multiple modes corresponding to plausible locations
(Fig.~\ref{fig:uncertainty}). PlaNet outperforms the Im2GPS
approach \cite{Hays08CVPR,Hays14MLEVI} that shares a similar goal. A
small-scale experiment shows that PlaNet even reaches superhuman
performance at the task of geolocating street view scenes.
Moreover, we show that the features learned by PlaNet can be used for image retrieval and achieve state-of-the-art results on the INRIA Holidays dataset \cite{Jegou08ECCV}.

Sometimes an image provides no useful cues: this is often the case
with portraits, or photos of pets and common foods.  However, we could
still make predictions about the location of such photos if we also
consider photos taken at roughly the same time either before or after
the query. To this end, we have extended PlaNet to work on groups
of photos by combining it with an LSTM approach.
This method yields a 50\% improvement over the single-image model when
applied to photo albums. The reason for this improvement is that LSTMs
learn to exploit temporal coherence in albums to correctly
geolocate even those photos that the single-image model would fail
to annotate confidently. For example, a photo of a croissant could be
taken anywhere in the world, but if it is in the same album as a photo
of the Eiffel Tower, the LSTM model will use this cue to
geolocate it to Paris.

\section{Related Work}
\label{sec:relwork}
% Retrieval: Flickr, Landmarks, Street View
%
%% Flickr + global features
Given a query photo, Im2GPS \cite{Hays08CVPR,Hays14MLEVI} retrieves
similar images from millions of geotagged Flickr photos and assigns
the location of the closest match to the query. Image distances are
computed using a combination of global image descriptors. Im2GPS
shows that with enough data, even this simple approach can achieve
surprisingly good results. We discuss Im2GPS in detail in
Sec.~\ref{sec:method}.

%% Global features + Aerial Imagery
%
%Deep representations for ground to aerial geolocalization (belongie)
Because photo coverage in rural areas is sparse,
\cite{Lin13CVPR,Lin15CVPR} make additional use of satellite aerial
imagery. \cite{Lin15CVPR} use CNNs to learn a joint embedding for
ground and aerial images and localize a query image by matching it
against a database of aerial images.
\cite{Workman15ICCV} take a similar approach and use a CNN to
transform ground-level features to the feature space of aerial images.

%% Local features
Image retrieval based on local features, bags-of-visual-words (BoVWs)
and inverted indices \cite{Nister06CVPR,Sivic03ICCV} has been shown to
be more accurate than global descriptors at matching buildings,
but requires more space and lacks the invariance to match \eg natural scenes or articulated objects. Most local feature based
approaches therefore focus on localization within cities, either based
on photos from photo sharing websites \cite{Cao15IJCV,Philbin07CVPR}
%
%% Street view matching
or street view
\cite{Baatz10ECCV,Chen11CVPR,Kim15ICCV,Knopp10ECCV,Schindler07CVPR,Zamir10ECCV,Zamir14PAMI}.
%
% Also in cities:
Skyline2GPS \cite{Ramalingam10IROS} also uses street view data, but
takes a unique approach that segments the skyline out of an image
captured by an upward-facing camera and matches it against a 3D model
of the city.

% Pose estimation
While matching against geotagged images can provide the rough location
of a query photo, some applications require the exact 6-dof camera
pose. \emph{Pose estimation} approaches achieve this goal using 3D
models reconstructed using structure-from-motion from internet photos.
A query image is localized by establishing correspondences between its
interest points and the points in the 3D model and solving the
resulting perspective-n-point (PnP) problem to obtain the camera
parameters
\cite{Li10ECCV,Li12ECCV,Sattler11ICCV}. Because matching the query
descriptors against the 3D model descriptors is expensive, some
approaches combine this technique with efficient image retrieval based
on inverted indices \cite{Cao15IJCV,Irschara09CVPR,Sattler12BMVC}.

% Landmark recognition
Instead of matching against a flat collection of photos, landmark
recognition systems
\cite{Avrithis10MM,Gammeter09ICCV,Johns11ICCV,Quack08CIVR,Zheng09CVPR}
build a database of \emph{landmark buildings} by clustering images
from internet photo collections. The landmarks in a query image are
recognized by retrieving matching database images and returning the
landmark associated with them.
% ... as classification
Instead of using image retrieval, \cite{Bergamo13CVPR,Li09CVPR} use
SVMs trained on BoVW of landmark clusters to decide which landmark is
shown in a query image. Instead of operating on image clusters,
\cite{Gronat13CVPR}, train one exemplar SVM for each image in a
dataset of street view images.

% Scene recognition
A task related to image geolocation is scene recognition, for
which the SUN database \cite{Xiao14IJCV} is an established benchmark.
The database consists of 131k images categorized into 908 scene
categories such as ``mountain``, ``cathedral`` or ``staircase``. The SUN
survey paper \cite{Xiao14IJCV} shows that Overfeat \cite{Sermanet14ICLR}, a CNN trained on
ImageNet \cite{Deng09CVPR} images, consistently outperforms other approaches, including
global descriptors like GIST and local descriptors like SIFT,
motivating our use of CNNs for image geolocation.

% Sequence geolocalization
In Sec.~\ref{sec:lstm}, we extend PlaNet to geolocate
sequences of images using LSTMs. Several previous approaches have also
realized the potential of exploiting temporal coherence to
geolocate images.
\cite{Chen11CVPRb,Li09CVPR} first cluster the photo collection into
landmarks and then learn to predict the sequence of landmarks in a
query photo sequence.  While \cite{Chen11CVPRb} train a Hidden Markov
Model (HMM) on a dataset of photo albums to learn popular tourist
routes, \cite{Li09CVPR} train a structured SVM that uses temporal
information as an additional feature.
Images2GPS \cite{Kalogerakis09ICCV} also trains an HMM, but instead of
landmarks, its classes are a set of geographical cells partitioning
the surface of the earth. This is similar to our approach, however we
use  a much finer discretization.

Instead of performing geolocation, \cite{Zhang15WACV} train a CNN on a
large collection of geotagged Flickr photos to predict geographical
attributes like ``population``, ``elevation`` or ``household
income``. \cite{Frey14WACV} cluster street view imagery to discover
latent scene types that are characteristic for certain geographical
areas and analyze how these types correlate with geopolitical boundaries.

In summary, most previous approaches to photo geolocation are
restricted to urban areas which are densely covered by street view
imagery and tourist photos. Exceptions are Im2GPS
\cite{Hays14MLEVI,Hays08CVPR} and
\cite{Lin13CVPR,Lin15CVPR,Workman15ICCV}, which make additional use of
satellite imagery. Prior work has shown that CNNs are well-suited for
scene classification \cite{Xiao14IJCV} and geographical attribute
prediction \cite{Zhang15WACV}, but to our knowledge ours is the first
method that directly takes a classification approach to geolocation
using CNNs.

\section{Image Geolocation with CNNs}
\label{sec:method}
We pose the task of image geolocation as a classification problem.
For this, we subdivide the earth into a set of geographical cells.
The input to our CNN are the image pixels and the
target output is a one-hot vector encoding the cell containing the
geotag of the image. Given a test image,
the output of this model is a probability
distribution over the world. The advantage of this formulation over a
regression from pixels to latitude/longitude coordinates is that the
model can express its uncertainty about an image by assigning each
cell a confidence that the image was taken there. In contrast, a
regression model would be forced to pinpoint a single location and
would have no natural way of expressing uncertainty about its
prediction, especially in the presence of multi-modal answers (as are
expected in this task).

\begin{figure}[t]
  \centering
  \includegraphics[width=.25\linewidth]{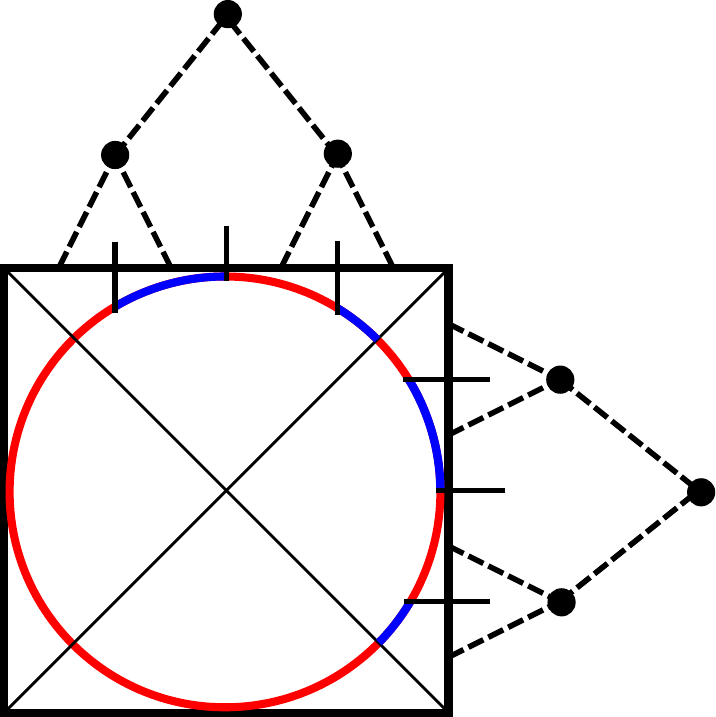}
  \caption{S2 cell quantization in 2D. The sides of the square are subdivided recursively and projected onto the circle.}
  \label{fig:s2}
\end{figure}

\PAR{Adaptive partitioning using S2 Cells.}
\begin{figure*}[t]
  \centering
  \includegraphics[width=.49\linewidth]{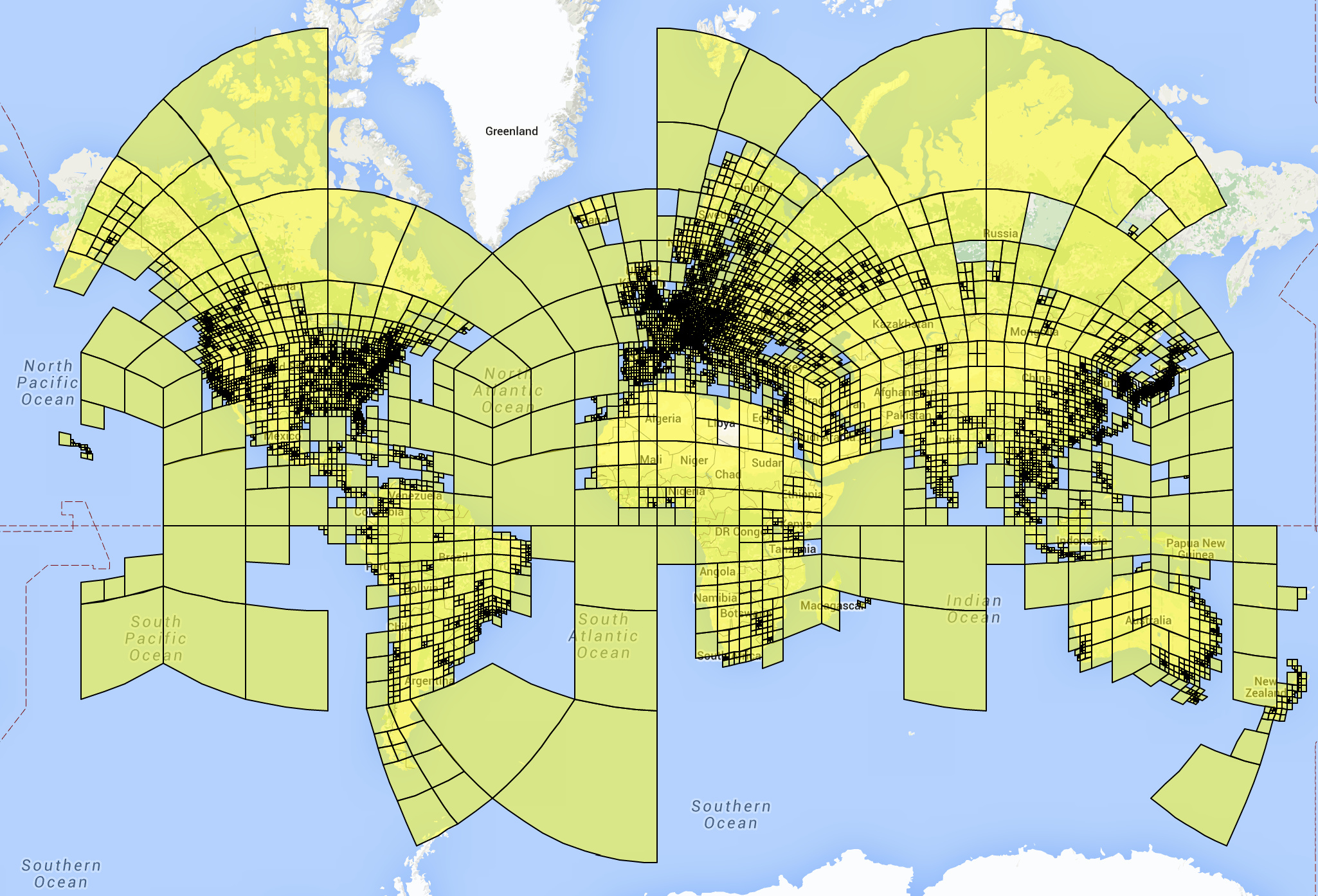}
  \includegraphics[width=.25\linewidth]{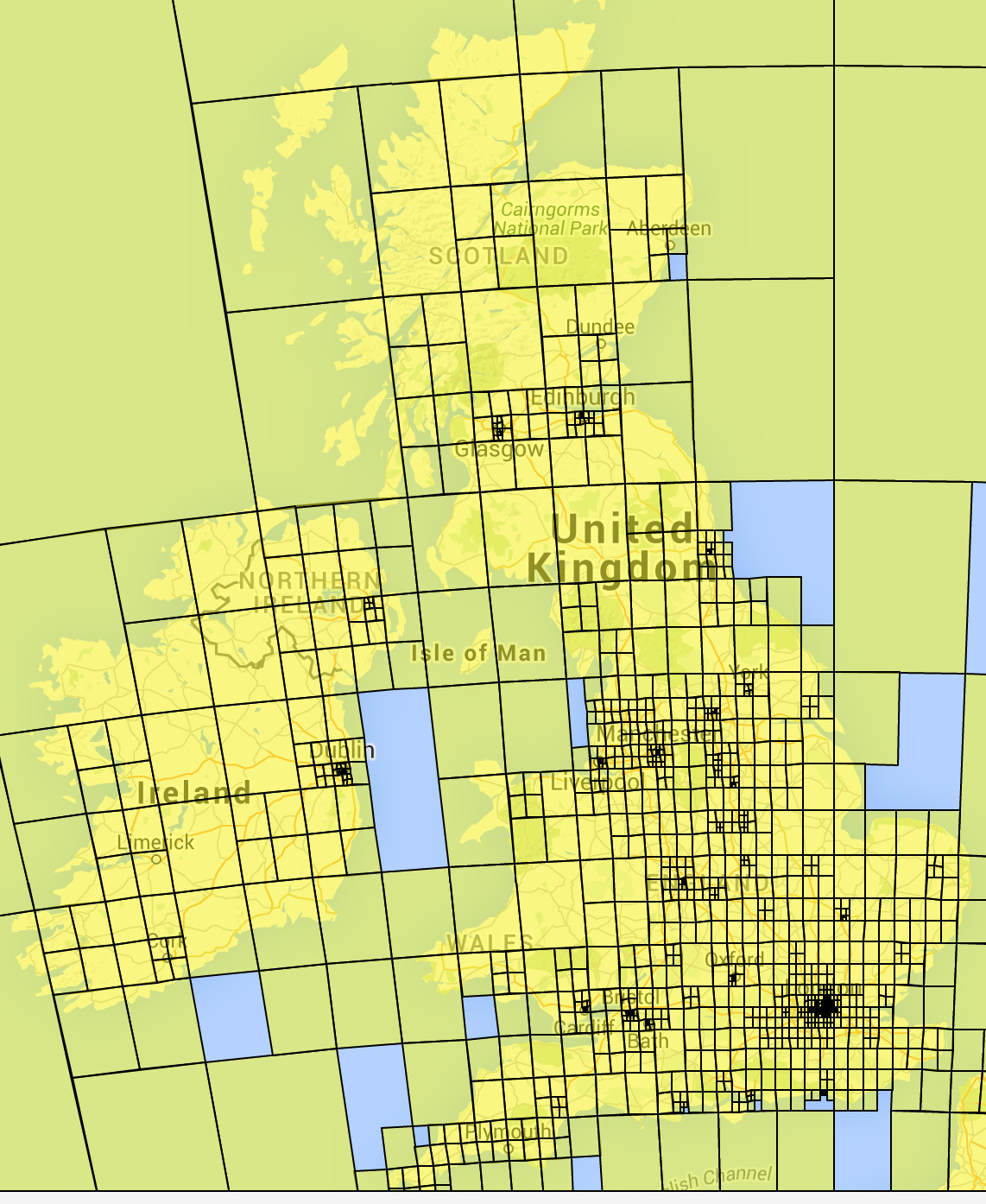}   \includegraphics[width=.25\linewidth]{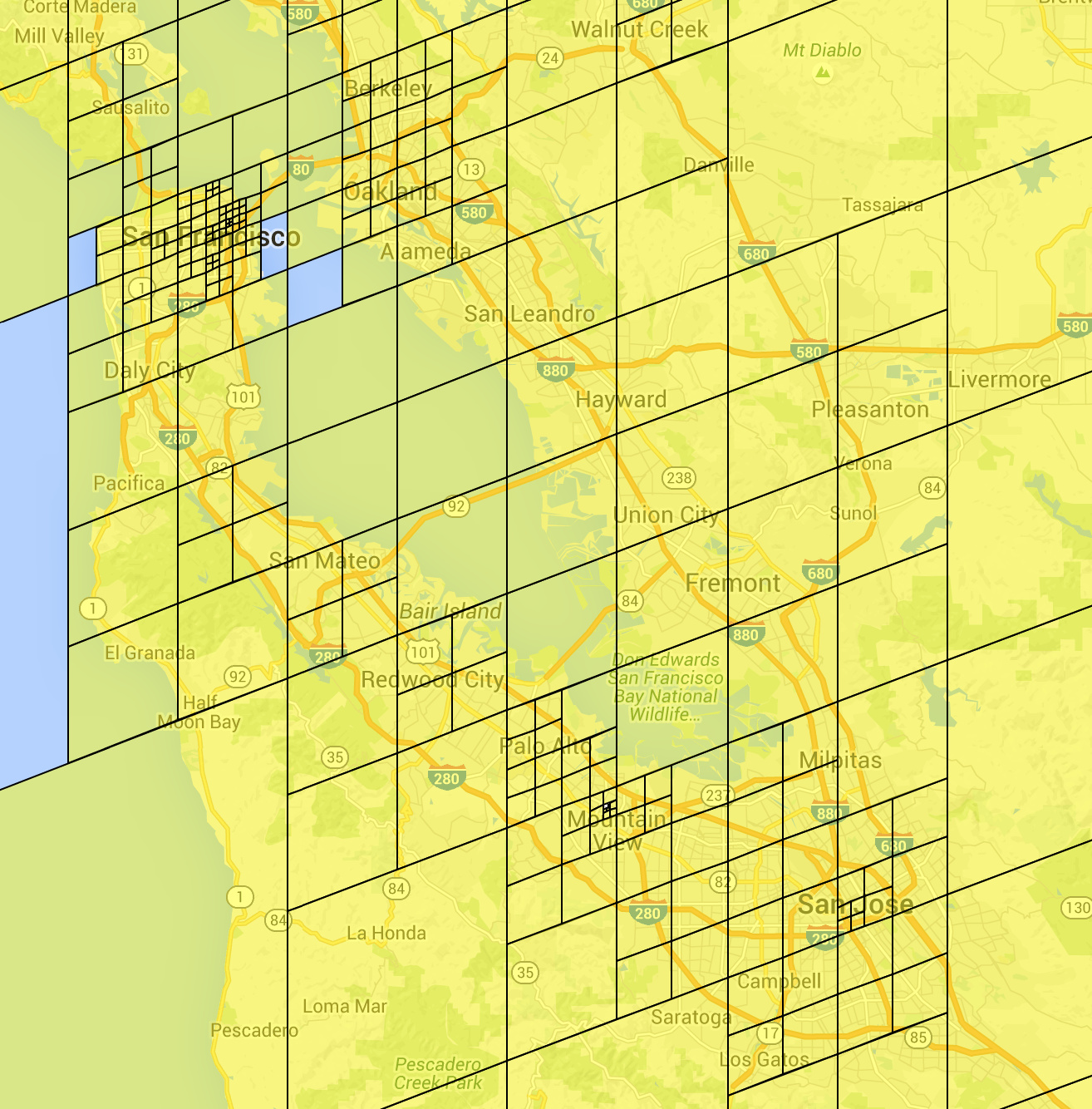}
  \caption{Left: Adaptive partitioning of the world into 26,263 S2 cells. Right: Detail views of Great Britain and Ireland and the San Francisco bay area.}
  \label{fig:s2_partitioning}
\end{figure*}
We use Google's open source S2 geometry
library\footnote{\href{https://code.google.com/p/s2-geometry-library/}{https://code.google.com/p/s2-geometry-library/}}\footnote{\href{https://docs.google.com/presentation/d/1Hl4KapfAENAOf4gv-pSngKwvS\_jwNVHRPZTTDzXXn6Q/view}{https://docs.google.com/presentation/d/1Hl4KapfAENAOf4gv-pSngKwvS\_jwNVHRPZTTDzXXn6Q/view}}
to partition the earth's surface into a set of non-overlapping cells
that define the classes of our model. The S2 library defines a
hierarchical partitioning of the surface of a sphere by projecting the
surfaces of an enclosing cube on it. The six sides of the cube are
subdivided hierarchically by six quad-trees. A node in a quad-tree
defines a region on the surface of the sphere called an S2 cell.
Fig.~\ref{fig:s2} illustrates this in 2D.
We chose this subdivision scheme over a simple subdivision of
latitude/longitude coordinates, because (i) lat/lon regions get
elongated near the poles while S2 cells keep a close-to-quadratic
shape, and (ii) S2 cells have mostly uniform size (the ratio between
the largest and smallest S2 cell is 2.08).

A naive approach to define a tiling of the earth would be to use all
S2 cells at a certain fixed depth in the hierarchy, resulting in a set
of roughly equally sized cells (see Fig.~\ref{fig:uncertainty}). However, this would produce a very
imbalanced class distribution since the geographical distribution of
photos has strong peaks in densely populated areas. We therefore
perform adaptive subdivision based on the photos' geotags: starting at the roots, we recursively
descend each quad-tree and subdivide cells until no cell contains more
than a certain fixed number $t_1$ of photos. This way, sparsely
populated areas are covered by larger cells and densely populated
areas are covered by finer cells. Then, we discard all cells
containing less than a minimum of $t_2$ photos. Therefore, PlaNet does not
cover areas where photos are very unlikely to be taken, such as oceans
or poles. We remove all images from the training set that are in
any of the discarded cells. This adaptive tiling has several
advantages over a uniform one: (i) training classes are more balanced,
(ii) it makes effective use of the parameter space because more model
capacity is spent on densely populated areas, (iii) the model can reach
up to street-level accuracy in city areas where cells are small.
Fig.~\ref{fig:s2_partitioning} shows the S2 partitioning for
our dataset.

\PAR{CNN training.}
We train a CNN based on the Inception
architecture \cite{Szegedy15CVPR} with batch
normalization \cite{Ioffe15ICML}. The SoftMax output layer has one
output for each S2 cells in the partitioning. We
set the target output value to $1.0$ at the index of the S2 cell the respective
training image belongs to and all other outputs to $0.0$. We initialize the weights of our model with random values and use the AdaGrad \cite{Duchy11JMLR} stochastic gradient descent with a learning rate of 0.045.

Our dataset consists of 126M photos with Exif geolocations
mined from all over the web. We applied very little filtering, only excluding
images that are non-photos (like diagrams, clip-art, etc.) and porn.
Our dataset is therefore extremely noisy, including indoor photos,
portraits, photos of pets, food, products and other photos not
indicative of location. Moreover, the Exif geolocations may be
incorrect by several hundred meters due to noise. We split the dataset into 91M training images and 34M validation images.

For the adaptive S2 cell partitioning (Sec.~\ref{sec:method}) we set $t_1=10,000$
and $t_2=50$. The
resulting partitioning consists of $26,263$ S2 cells
(Fig.~\ref{fig:s2_partitioning}). Our Inception model has a total of
97,321,048 parameters.
We train the model for 2.5 months on 200 CPU cores using the DistBelief framework \cite{Dean12NIPS} until the accuracy on the validation set converges. The long training time
is due to the large variety of the training data and the large
number of classes.

We ensure that none of the test sets we use in this paper have any (near-) duplicate images in the training set. For this, we use a CNN trained on near-duplicate images to compute a binary embedding for each training and test image and then remove test images whose Hamming distance to a training image is below an aggressively chosen threshold. 

\PAR{Geolocation accuracy.}
\begin{figure}[t]
  \centering
 \includegraphics[width=\linewidth]{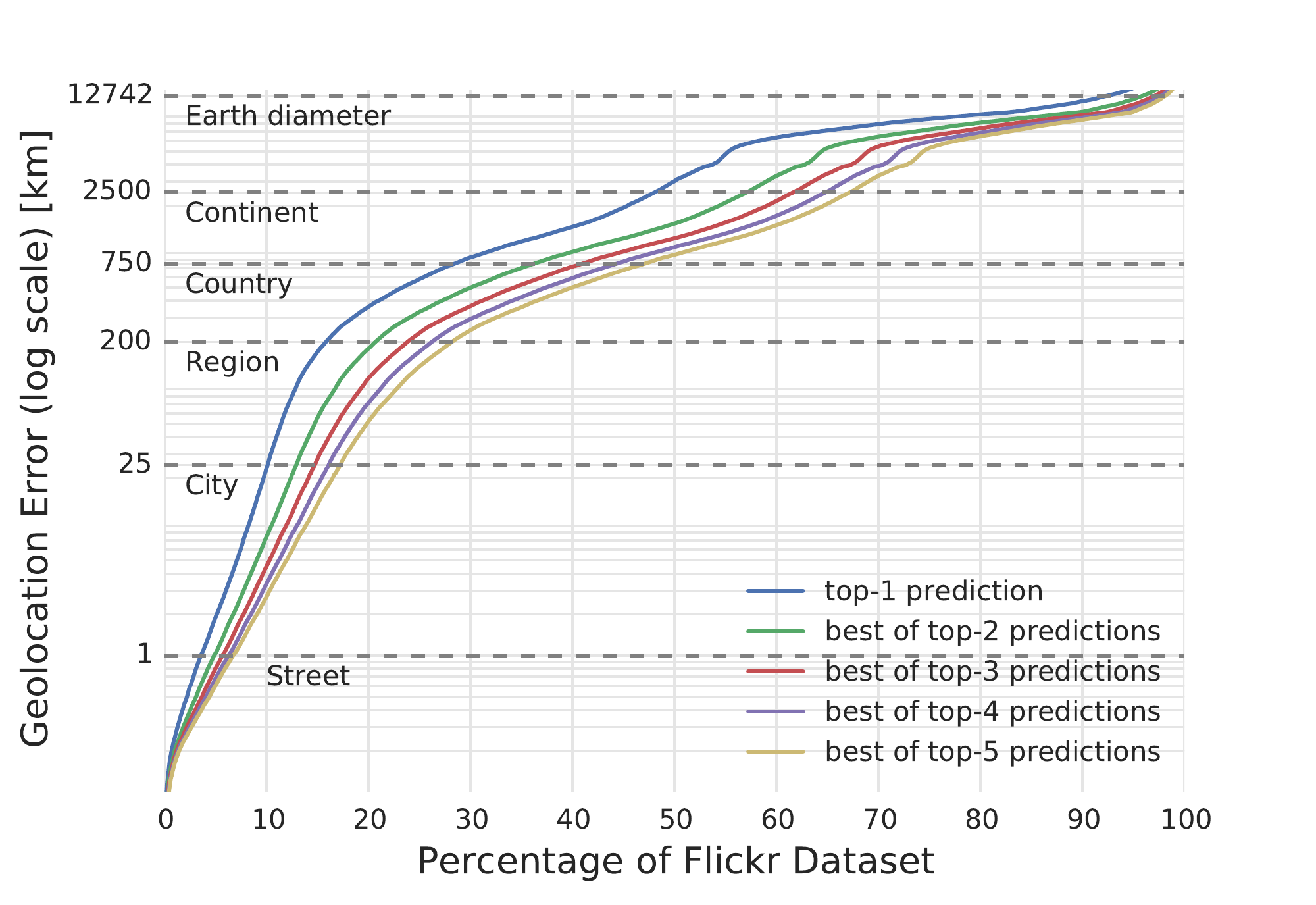}
 \caption{Geolocation accuracy of the top-k most confident predictions on 2.3M Flickr photos. (Lower right is best.)}
  \label{fig:flickr230m_plot}
\end{figure}
To quantitatively measure the localization accuracy of the model, we collected a dataset of 2.3M geotagged Flickr photos from across the world. Other than selecting geotagged images with 1 to 5 textual tags, we did not apply any filtering. Therefore, most of the images have little to no cues about their location.

To compute localization error, we run inference and compute the distance between the center of the predicted S2 cell to the original location of the photo. We note that this error measure is pessimistic, because even if the ground truth location is within the predicted cell, the error can still be large depending on the cell size.
Fig.~\ref{fig:flickr230m_plot} shows what fraction of this dataset was localized with a certain geographical distance of the ground truth locations. The blue curve shows the performance for the most confident prediction, and the other curves show the performance for the best of the top-\{2,3,4,5\} predictions per image. Following \cite{Hays14MLEVI}, we added approximate geographical scales of streets, cities, regions, countries and continents.
Despite the difficulty of the data, PlaNet is able to localize 3.6\% of the images at street-level accuracy and 10.1\% at city-level accuracy. 28.4\% of the photos are correctly localized at country level and 48.0\% at continent level. When considering the best of the top-5 predictions, the model localizes roughly twice as many images correctly at street, city, region and country level.  

\PAR{Qualitative Results.}
An important advantage of our localization-as-classification paradigm
is that the model output is a probability distribution over the globe.
This way, even if an image can not be confidently localized, the model
outputs confidences for possible locations. To illustrate this, we
trained a smaller model
using only S2 cells at level 4 in the S2
hierarchy, resulting in a total of only 354 S2 cells.
Fig.~\ref{fig:uncertainty} shows the predictions of this model for
test images with different levels of geographical ambiguity.

\begin{figure}[t]
  \centering
  \includegraphics[width=\linewidth]{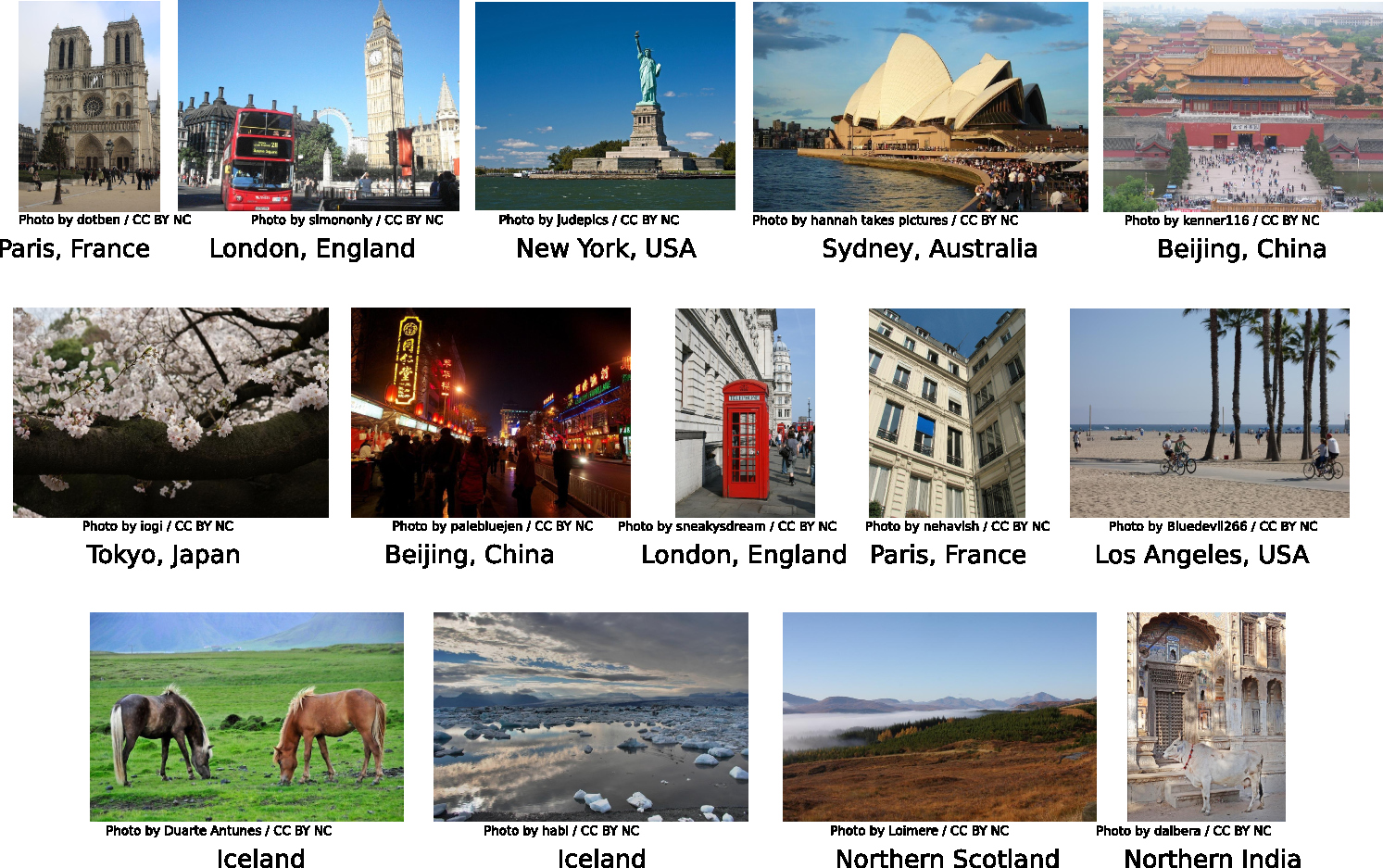}
  \caption{Examples of images PlaNet localizes correctly. Our
model is capable of localizing photos of famous landmarks (top row),
but often yields surprisingly accurate results for images with more
subtle geographical cues. The model learns to recognize locally
typical landscapes, objects, architectural styles and even plants and
animals.}
  \label{fig:correctly_localized}
\end{figure}

Fig.~\ref{fig:correctly_localized} shows examples of the
different types of images PlaNet can localize.
landmarks, which can also be recognized by landmark recognition
engines \cite{Avrithis10MM,Quack08CIVR,Zheng09CVPR}, PlaNet can
often correctly localize street scenes, landscapes, buildings of
characteristic architecture, locally typical objects like red phone
booths, and even some plants and animals.
\begin{figure}[t]
  \centering
  \includegraphics[width=\linewidth]{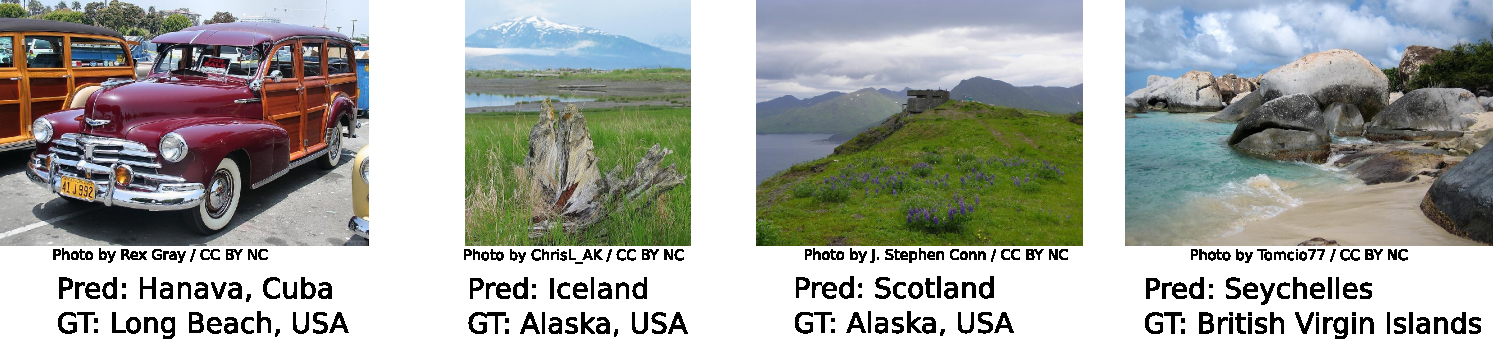}
  \caption{Examples of incorrectly localized images.}
  \label{fig:incorrectly_localized}
\end{figure}
Fig.~\ref{fig:incorrectly_localized} shows some failure modes.
Misclassifications often occur due to ambiguity, \eg, because certain
landscapes or objects occur in multiple places, or are more typical
for a certain place than the one the photo was taken (\eg, the kind of
car in the first image is most typically found in Cuba).

\begin{figure}[t]
  \centering
  \includegraphics[width=\linewidth]{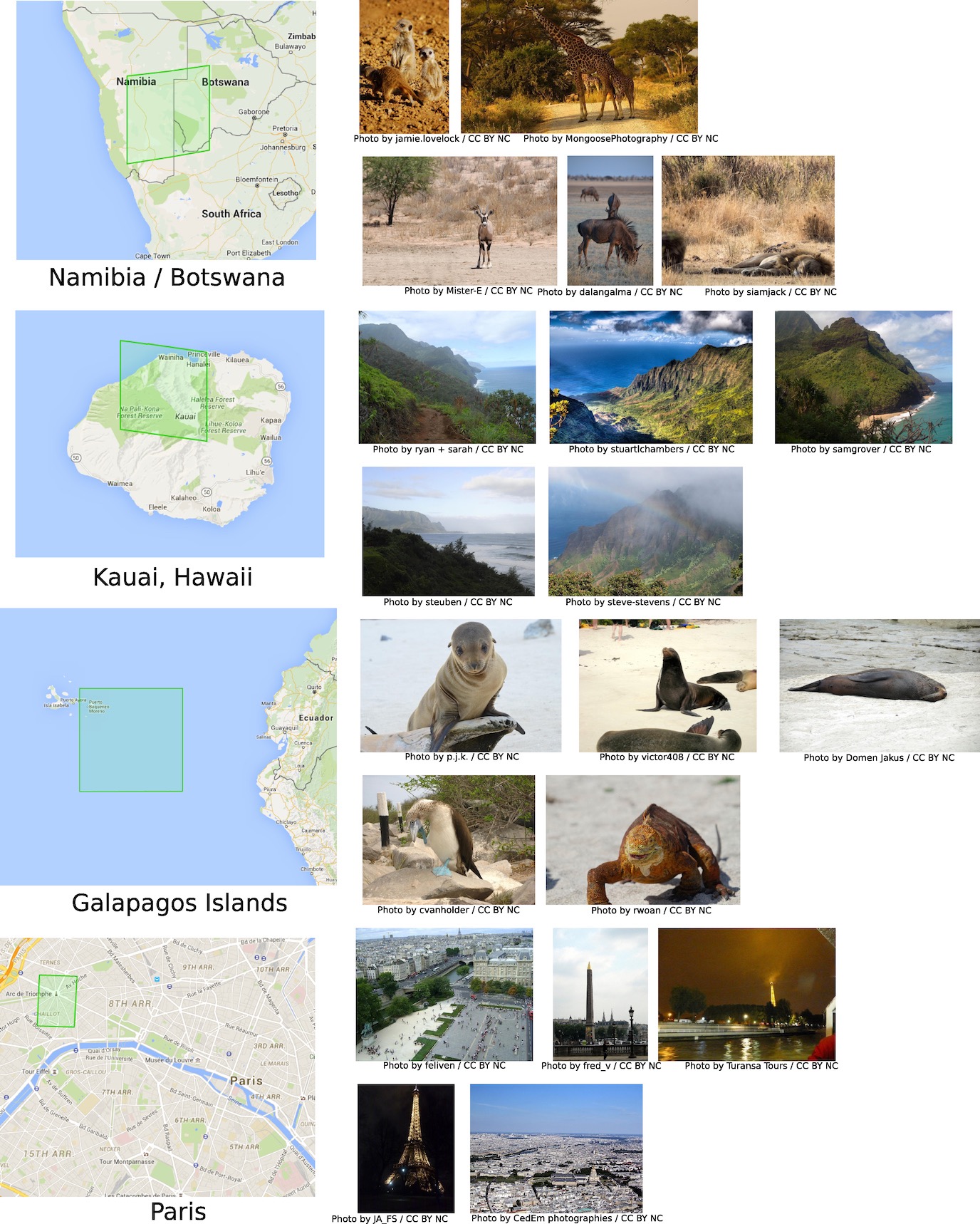}
  \caption{The top-5 most confident images from the Flickr dataset for
the S2 cells on the left, showing the diverse visual representation of
places that PlaNet learns.}
  \label{fig:cell_iconics}
\end{figure}

To give a visual impression of the representations PlaNet has
learned for individual S2 cells, Fig.~\ref{fig:cell_iconics} shows the test images
that the model assigns to a given cell with the highest confidence. The model learns a very
diverse representation of a single cell, containing the different
landmarks, landscapes, or animals that a typical for a specific
region.

\PAR{Comparison to Im2GPS.}
\begin{table}[t]
  \centering
  \footnotesize
    \setlength\tabcolsep{3pt}
   \begin{tabular}{lrrrrr}
     \hline
    \textbf{Method} & \textbf{\shortstack{Street\\1 km}} & \textbf{\shortstack{City\\25 km}} & \textbf{\shortstack{Region\\200 km}} & \textbf{\shortstack{Country\\750 km}} & \textbf{\shortstack{Continent\\2500 km}} \\
    \hline
 Im2GPS (orig) \cite{Hays08CVPR} & & 12.0\% & 15.0\% & 23.0\% & 47.0\% \\
Im2GPS (new) \cite{Hays14MLEVI} & 2.5\% & 21.9\% & 32.1\% & 35.4\% & 51.9\% \\
PlaNet & \textbf{8.4\%} & \textbf{24.5\%} & \textbf{37.6\%} & \textbf{53.6\%} & \textbf{71.3\%} \\
\hline
 \end{tabular}
 \caption{Comparison of PlaNet with Im2GPS. Percentages are the fraction of images from the Im2GPS test set that were localized within the given radius. (Numbers for the original Im2GPS are approximate as they were extracted from a plot in the paper.)}
 \label{tab:im2s2_vs_im2gps}
\end{table}
As discussed in Sec.~\ref{sec:relwork}, most image-based localization
approaches focus on photos taken inside cities. One of the few
approaches that, like ours, aims at geolocating arbitrary photos is
Im2GPS~\cite{Hays08CVPR,Hays14MLEVI}. However, instead of
classification, Im2GPS is based on nearest neighbor matching. The
original Im2GPS approach~\cite{Hays08CVPR} matches the query image
against a database of 6.5M Flickr images and returns the geolocation
of the closest matching image. Images are represented by a combination
of six different global image descriptors.
A recent extension of Im2GPS~\cite{Hays14MLEVI} uses both an improved
image representation and a more sophisticated localization technique.
It estimates a per-pixel probability of being ``ground``, ``vertical``,
``sky``, or ``porous`` and computes color and texture histograms for each
of these classes. Additionally, bag-of-visual-word vectors of length
1k and 50k based on SIFT features are computed for each image. The
geolocation of a query is estimated by retrieving nearest neighbors,
geo-clustering them with mean shift, training 1-vs-all SVMs for each
resulting cluster, and finally picking the average GPS coordinate of
the cluster whose SVM gives the query image the highest positive
score.

We evaluate PlaNet on the Im2GPS test dataset \cite{Hays08CVPR}
that consists of 237 geotagged photos from Flickr, curated such that
most photos contain at least a few geographical cues. 
Tab.~\ref{tab:im2s2_vs_im2gps} compares the performance of
PlaNet and both versions of Im2GPS. The new version is a
significant improvement over the old one. However, PlaNet
outperforms even the new version with a considerable margin. In
particular, PlaNet localizes 236\% more images accurately at
street level. The gap narrows at coarser scales, but even at country level
PlaNet still localizes 51\% more images accurately.

A caveat of our evaluation is that PlaNet was trained on 14x
more data than Im2GPS uses, which certainly gives PlaNet an
advantage. However, note that because Im2GPS performs a nearest
neighbor search, its runtime grows with the number of images while CNN
evaluation speed is independent of the amount of training data. Since
the Im2GPS feature vectors have a dimensionality of 100,000, Im2GPS
would require 8.5GB to represent our corpus of 91M training examples
(assuming one byte per descriptor dimension, not counting the space
required for a search index). In contrast, our model uses only 377 MB,
which even fits into the memory of a smartphone.

\PAR{Comparison to human performance.}
\begin{figure}
  \includegraphics[width=\linewidth]{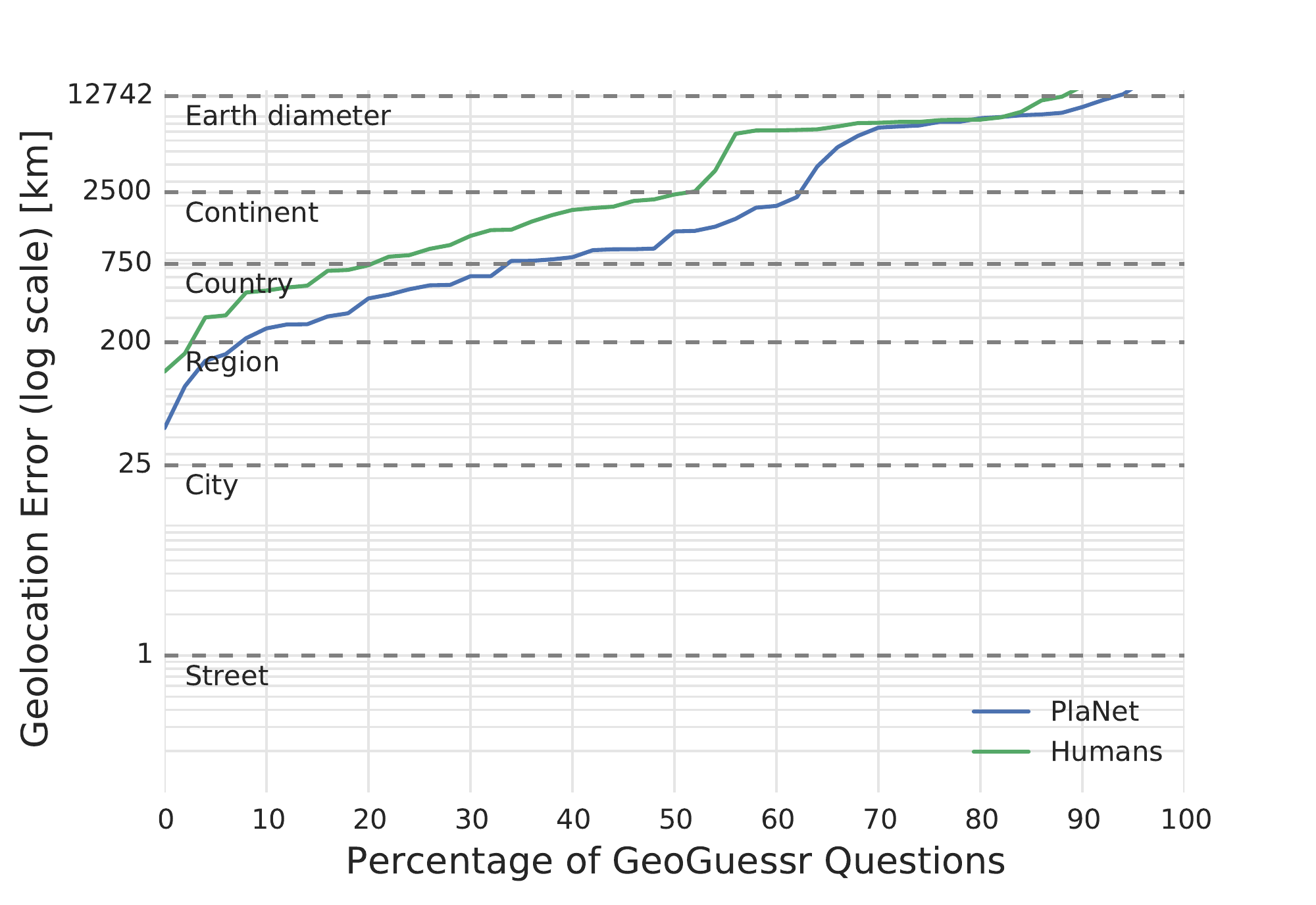}
  \caption{Geolocation error of PlaNet \vs humans.}
  \label{fig:geoguessr_plot}
\end{figure}
To find out how PlaNet compares with human intuition, we let it compete against 10
well-traveled human subjects in a game of Geoguessr
(\href{http://www.geoguessr.com}{www.geoguessr.com}). Geoguessr
presents the player with a random street view panorama (sampled from
all street view panoramas across the world) and asks them to place a
marker on a map at the location the panorama was captured. Players are
allowed to pan and zoom in the panorama, but may not navigate to
adjacent panoramas. The map is zoomable, so the location can be
specified as precisely as the player wants. For this experiment, we
used the game's ``challenge mode'' where two players are shown the
same set of 5 panoramas. We entered the PlaNet guesses manually
by taking a screenshot of the view presented by the game, running
inference on it and entering the center of the highest confidence S2
cell as the guess of the PlaNet player. For a fair comparison,
we did not allow the human subjects to pan and zoom, so they did not
use more information than we gave to the model. For each subject, we
used a different set of panoramas, so humans and PlaNet played a
total of 50 different rounds.

\begin{figure*}[t]
  \centering
  \def \s {.195}
  \setlength\tabcolsep{1pt}
  \begin{tabular}{ccccc}
    \includegraphics[width=\s\linewidth]{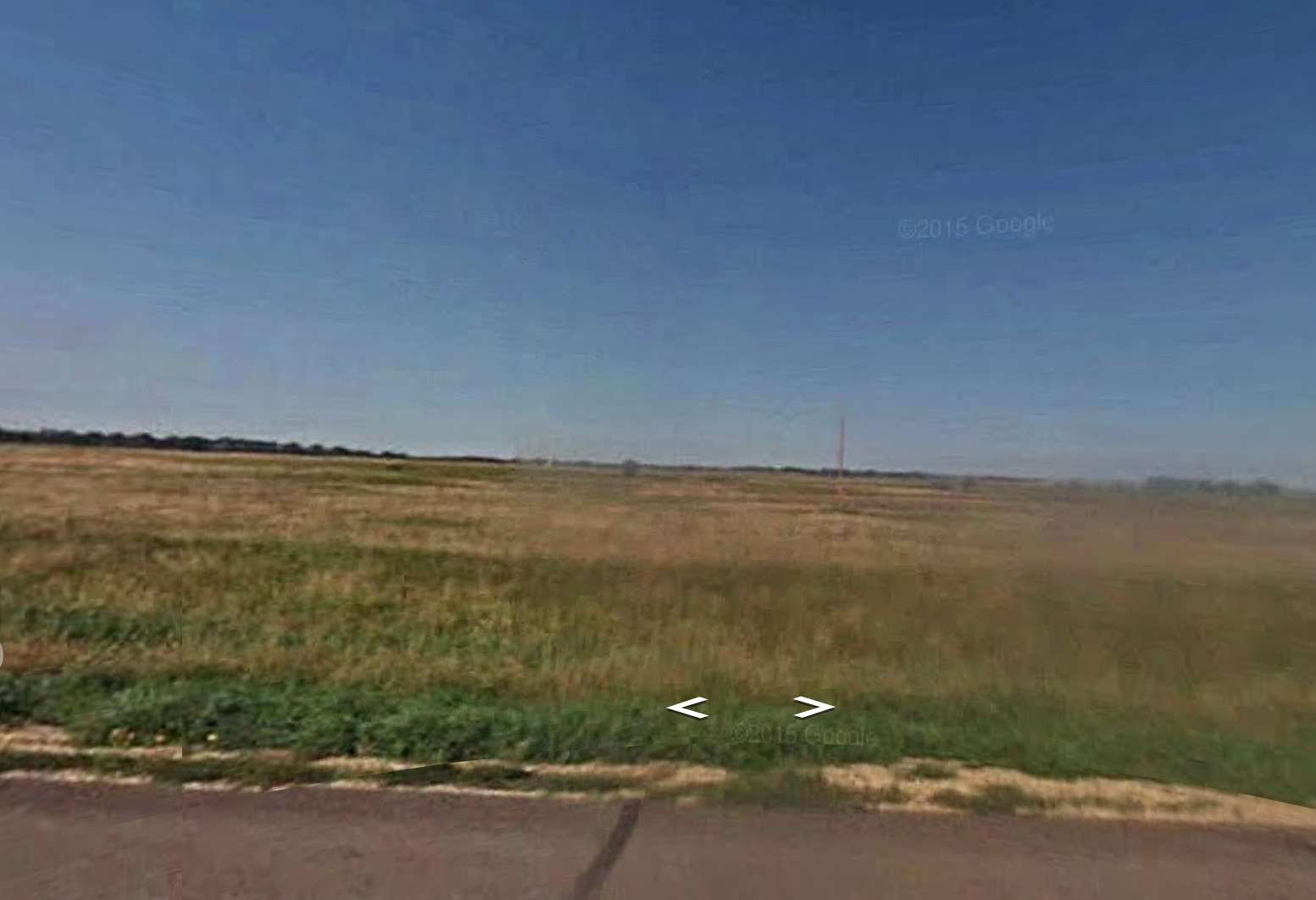} &
    \includegraphics[width=\s\linewidth]{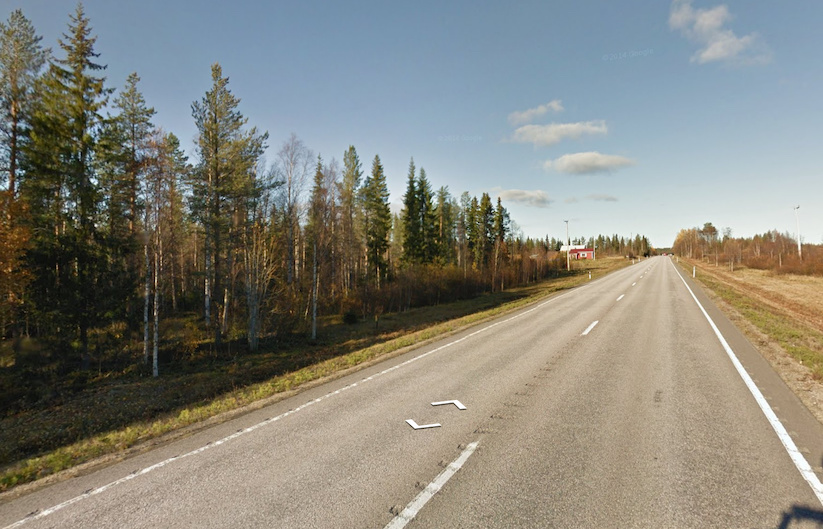} &
    \includegraphics[width=\s\linewidth]{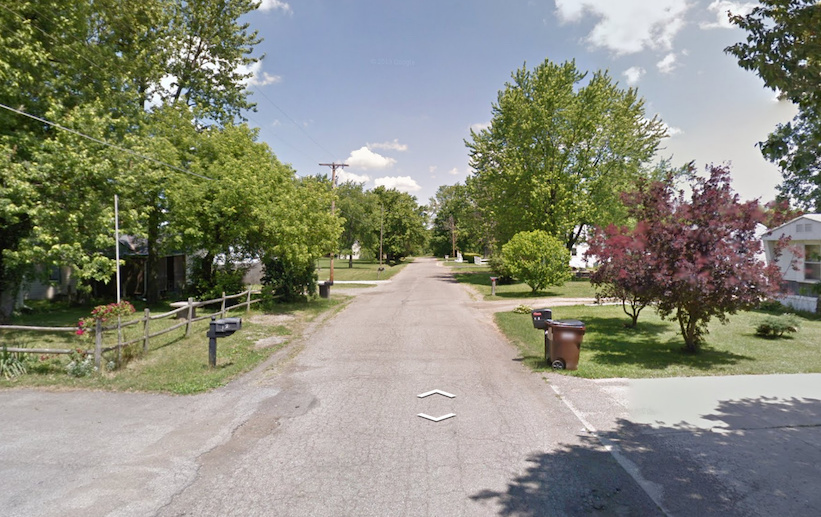} &
    \includegraphics[width=\s\linewidth]{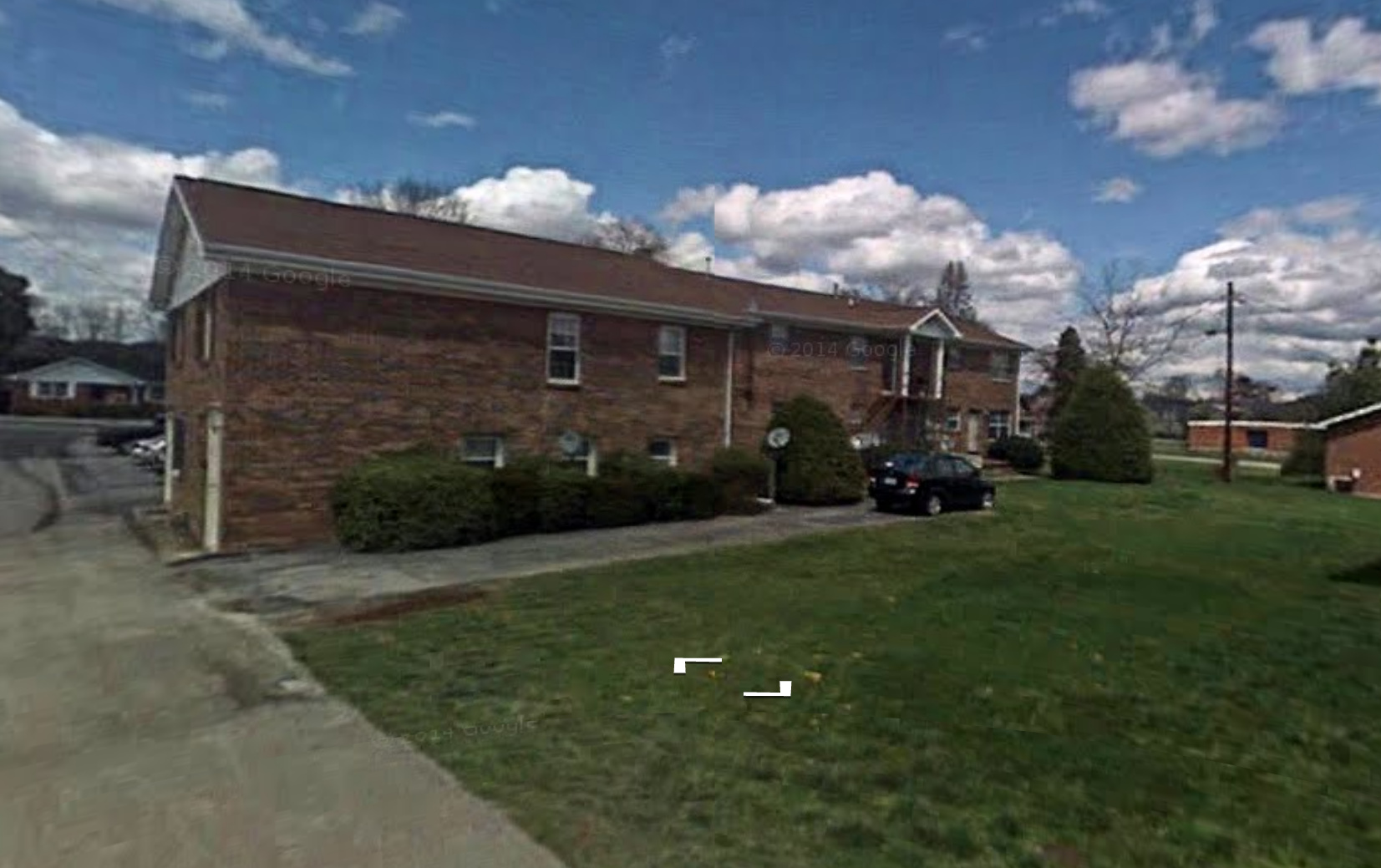} &
    \includegraphics[width=\s\linewidth]{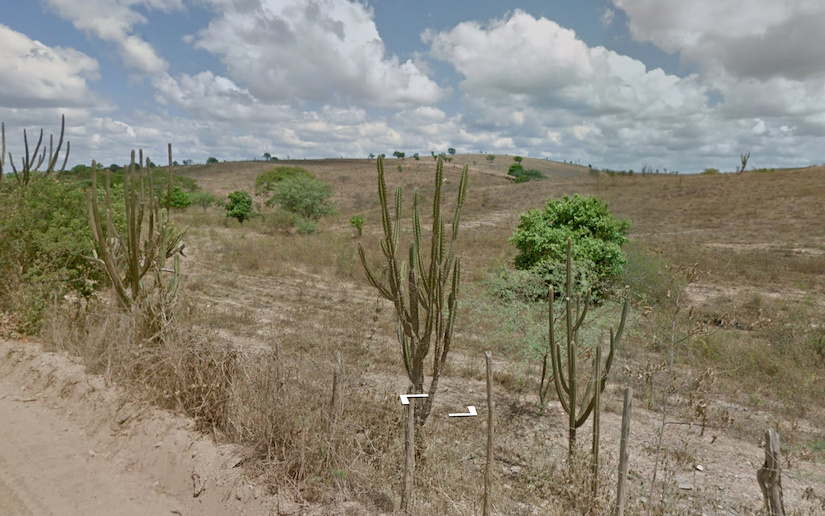} \\
    \includegraphics[width=\s\linewidth]{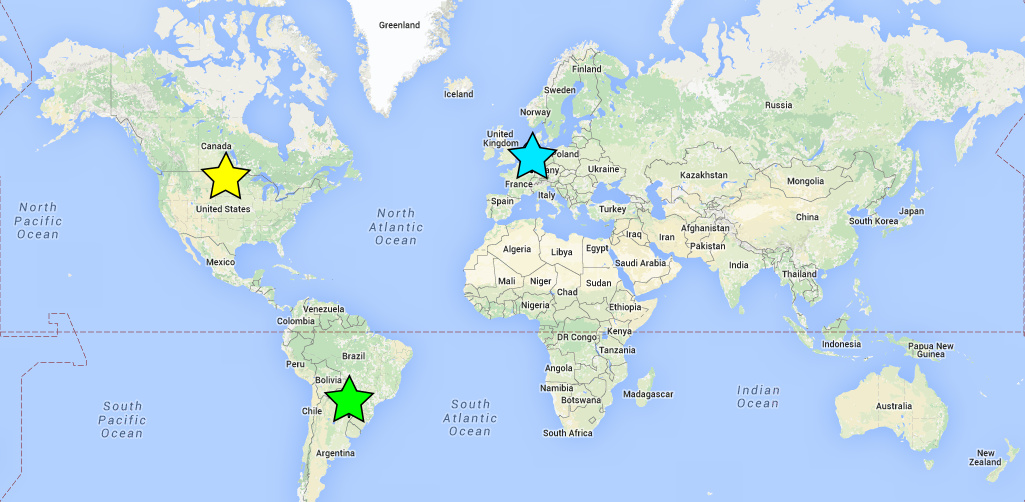} &
    \includegraphics[width=\s\linewidth]{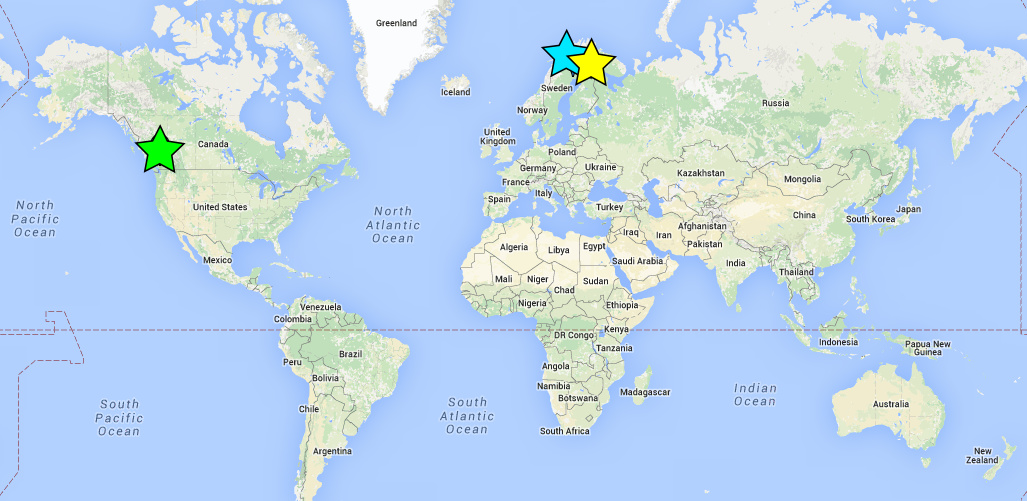} &
    \includegraphics[width=\s\linewidth]{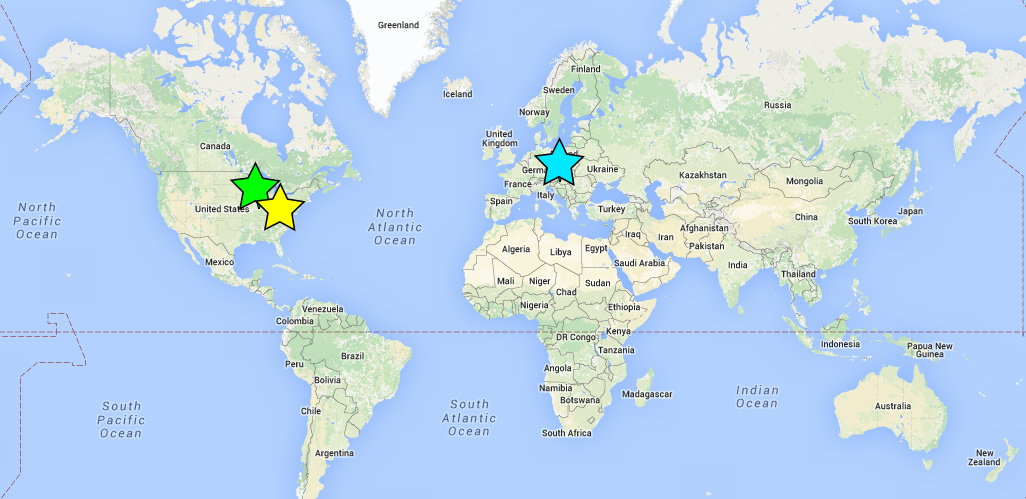} &
    \includegraphics[width=\s\linewidth]{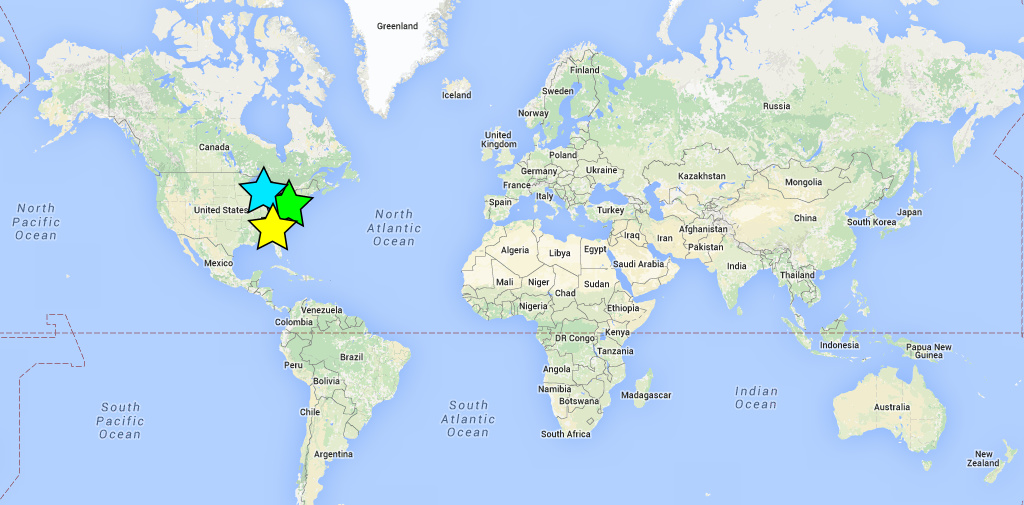} &
    \includegraphics[width=\s\linewidth]{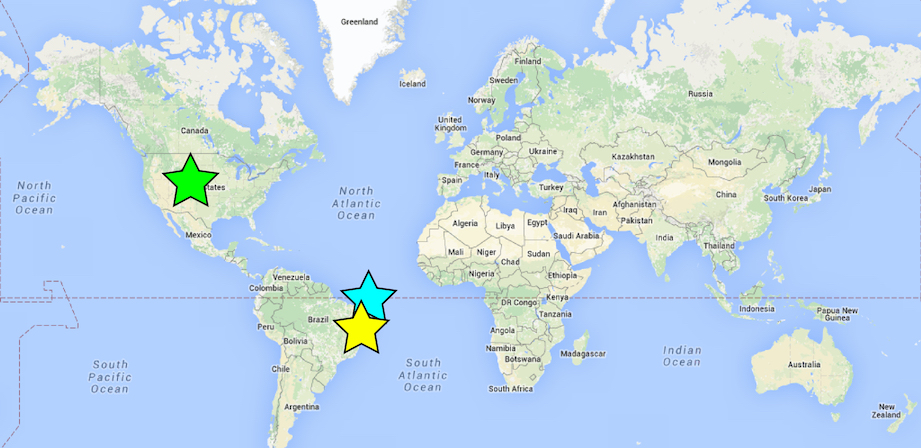}
  \end{tabular}
  \caption{Top: GeoGuessr panorama, Bottom: Ground truth location (yellow), human guess (green), PlaNet guess (blue).}
  \label{fig:geoguessr_examples}
\end{figure*}

In total, PlaNet won 28 of the 50 rounds with a median
localization error of 1131.7 km, while the median human localization
error was 2320.75 km. Fig.~\ref{fig:geoguessr_plot} shows what
percentage of panoramas were localized within which distance by humans
and PlaNet respectively. Neither humans nor PlaNet were
able to localize photos below street or city level, showing that this
task was even harder than the Flickr dataset and the Im2GPS dataset.
Fig.~\ref{fig:geoguessr_examples} shows some example panoramas from
the game together with the guessed locations. Most panoramas were
taken in rural areas containing little to no geographical cues.

When asked what cues they used, human subjects said they looked for
any type of signs, the types of vegetation, the architectural style,
the color of lane markings and the direction of traffic on the street.
Furthermore, humans knew that street view is not available in certain
countries such as China allowing them to further narrow down their
guesses. One would expect that these cues, especially street signs,
together with world knowledge and common sense should give humans an
unfair advantage over PlaNet, which was trained solely on image
pixels and geolocations. Yet, PlaNet was able to outperform
humans by a considerable margin. For example, PlaNet localized
17 panoramas at country granularity (750 km) while humans only
localized 11 panoramas within this radius. We think  PlaNet has
an advantage over humans because it has seen many more places than any
human can ever visit and has learned subtle cues of different scenes
that are even hard for a well-traveled human to distinguish.

\PAR{Features for image retrieval.}
A recent study \cite{Razavian14DVW} showed that the activations of Overfeat \cite{Sermanet14ICLR}, a CNN trained on ImageNet \cite{Deng09CVPR} can serve as powerful features for several computer vision tasks, including image retrieval. Since PlaNet was trained for location recognition, its features should be well-suited for image retrieval of tourist photos.
To test this, we evaluate the PlaNet features on the INRIA Holidays dataset \cite{Jegou08ECCV}, consisting of 1,491 personal holiday photos, including landmarks, cities and natural scenes.
We extract image embeddings from the final layer below the SoftMax layer (a 2048-dim.\ vector) and rank images by the Euclidean distance between their embedding vectors.  As can be seen in Tab.~\ref{tab:retrieval_results}, the PlaNet features outperform the Overfeat features. This is expected since our training data is more similar to the photos from the Holidays dataset than ImageNet. The same observation was made by \cite{Babenko14ECCV} who found that re-training on a landmark dataset improves retrieval performance of CNN features compared to those from a model trained on ImageNet. Their features even outperform ours, which is likely because \cite{Babenko14ECCV} use a carefully crafted landmark dataset for re-training while we applied only minimal filtering.
Using the \emph{spatial search} and \emph{augmentation} techniques described in \cite{Razavian14DVW}, PlaNet even outperforms state-of-the-art local feature based image retrieval approaches on the Holidays dataset.
We note that the Euclidean distance between these image embeddings is not necessarily meaningful as PlaNet was trained for classification. We expect Euclidean embeddings trained for image retrieval using a triplet loss \cite{Wang14CVPR} to deliver even higher mAP.

\begin{table}[t]
  \centering
  \footnotesize
  \begin{tabular}{lr}
    \hline
    \textbf{Method} & \textbf{Holidays mAP} \\
    \hline
    BoVW & 57.2 \cite{Jegou10IJCV} \\
    Hamming Embedding & 77.5 \cite{Jegou10IJCV} \\
    Fine Vocabulary & 74.9 \cite{Mikulik10ECCV} \\
    ASMK+MA & 82.2  \cite{Tolias13ICCV} \\
    \hline
    GIST & 37.6 \cite{Douze09CIVR} \\
    Overfeat & 64.2 \cite{Razavian14DVW} \\
    Overfeat+aug+ss & 84.3 \cite{Razavian14DVW} \\
    AlexNet+LM Retraining & 79.3  \cite{Babenko14ECCV} \\
    \hline
    PlaNet (this work) & 73.3 \\
    PlaNet+aug+ss & \textbf{89.9} \\
    \hline
  \end{tabular}
  \caption{Image retrieval mAP using PlaNet features compared to other methods.}
  \label{tab:retrieval_results}
\end{table}

\PAR{Model analysis.}
\begin{table}[t]
  \centering
  \footnotesize
    \setlength\tabcolsep{3pt}
   \begin{tabular}{lrrrrr}
     \hline
    \textbf{Method} & \textbf{\shortstack{Manmade\\Landmark}} & \textbf{\shortstack{Natural\\Landmark}} & \textbf{\shortstack{City\\Scene}} & \textbf{\shortstack{Natural\\Scene}} & \textbf{Animal} \\
    \hline
 Im2GPS (new) & 61.1 & 37.4 & 3375.3 & 5701.3 & 6528.0 \\
PlaNet & 74.5 & 61.0 & 212.6 & 1803.3 & 1400.0 \\
\hline
 \end{tabular}
 \caption{Median localization error (km) by image category on the Im2GPS test set. \emph{Manmade Landmark} are landmark buildings like the Eiffel Tower, \emph{Natural Landmark} are geographical features like the Grand Canyon, \emph{City Scene} and \emph{Natural Scene} are photos taken in cities or in nature not showing any distinctive landmarks, and \emph{Animal} are photos of individual animals.}
 \label{tab:categories}
\end{table}
For a deeper analysis of PlaNet's performance we manually classified the images of the Im2GPS test set into different scene types. Tab.~\ref{tab:categories} shows the median per-category error of PlaNet and Im2GPS. The results show that PlaNet's location discretization hurts its accuracy when pinpointing the locations of landmarks. However, PlaNet's clear strength is scenes, especially city scenes, which give it the overall advantage.

\begin{figure*}[t]
  \centering
  \subfloat[\label{fig:heatmap_grand_canyon}]{\includegraphics[width=.3\linewidth]{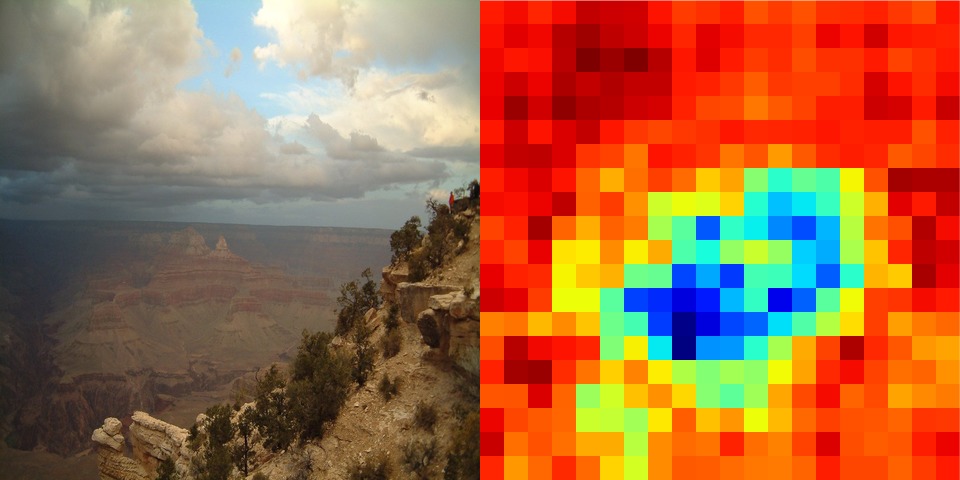}} \vspace{1pt}
\subfloat[\label{fig:heatmap_norway}]{\includegraphics[width=.3\linewidth]{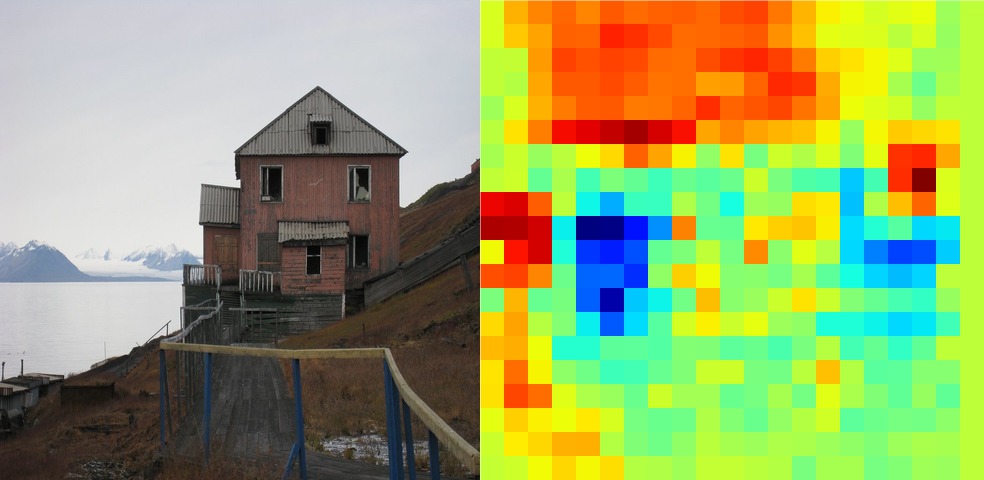}} \vspace{1pt}
 \subfloat[\label{fig:heatmap_shanghai}]{\includegraphics[width=.3\linewidth]{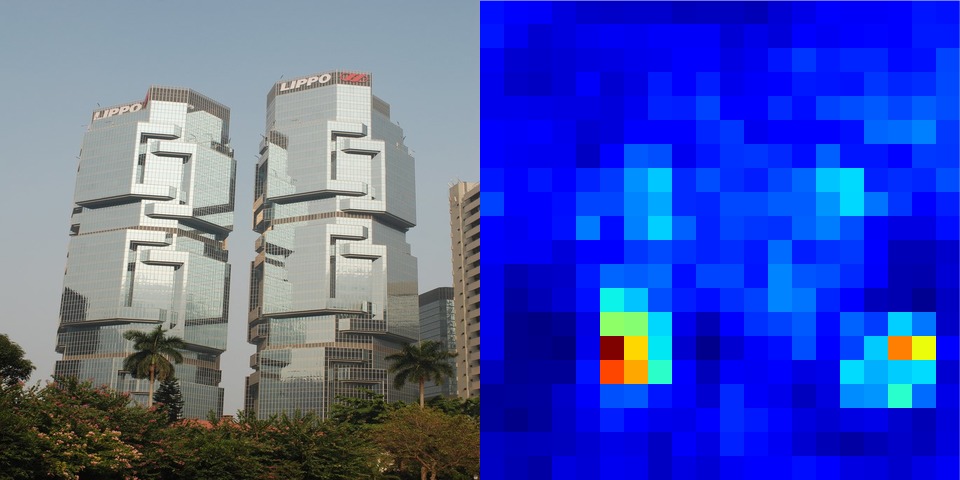}} \vspace{1pt}

  \caption{Left: Input image, right: Heatmap of the probability of the correct class when sliding an occluding window over the image as in \cite{Zeiler14ECCV}. (a) Grand Canyon. Occluding the distinctive mountain formation makes the confidence in the correct location drop the most. (b) Norway. While the house in the foreground is fairly generic, the snowy mountain range on the left is the most important cue. (c) Shanghai. Confidence in the correct location increases if the palm trees in the foreground are covered since they are not common in Shanghai.}
  \label{fig:heatmaps}
\end{figure*}
To analyze which parts of the input image are most important for the classifier's decision, we employ a method introduced by Zeiler \etal \cite{Zeiler14ECCV}. We plot an activation map where the value of each pixel is the classifier's confidence in the ground truth geolocation if the corresponding part of the image is occluded by a gray box (Fig.~\ref{fig:heatmaps}).

\section{Sequence Geolocation with LSTMs}
\label{sec:lstm}
While PlaNet is capable of localizing a large variety of images,
many images are ambiguous or do not contain enough information that
would allow to localize them. However we can exploit the fact that
photos naturally occur in sequences, \eg, photo albums, that have a high
geographical correlation. Intuitively, if we can confidently localize
some of the photos in an album, we can use this information to also
localize the photos with uncertain location. Assigning each photo in
an album a location is a sequence-to-sequence problem which requires a
model that accumulates a state from previously seen examples and makes
the decision for the current example based on both the state and the
current example. Therefore, long-short term memory (LSTM)
architectures \cite{Hochreiter97NC} seem like a good fit for this
task. We now explore how to address the problem of predicting photo
sequence geolocations using LSTMs.

\PAR{Training Data.}
For this task, we collected a dataset of 29.7M public photo albums
with geotags from Google+, which we split into 23.5M
training albums (490M images) and 6.2M testing albums (126M)
images. We use the S2 quantization scheme from the previous section to
assign labels to the images.

\PAR{Model architecture.}
\begin{figure*}[t]
  \centering
%.23
  \subfloat[\label{fig:lstm_model1}]{\includegraphics[width=.16\linewidth]{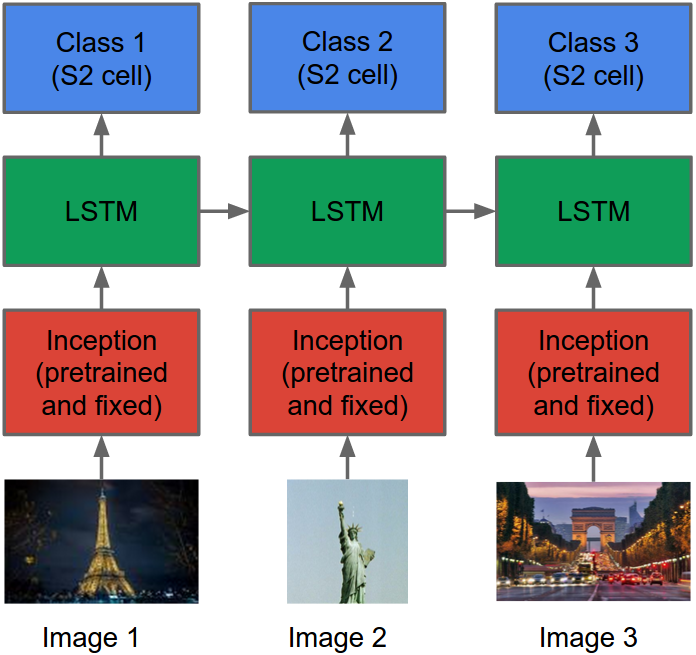}}
  \hspace{8pt}
%.23
\subfloat[\label{fig:lstm_offset}]{\includegraphics[width=.25\linewidth]{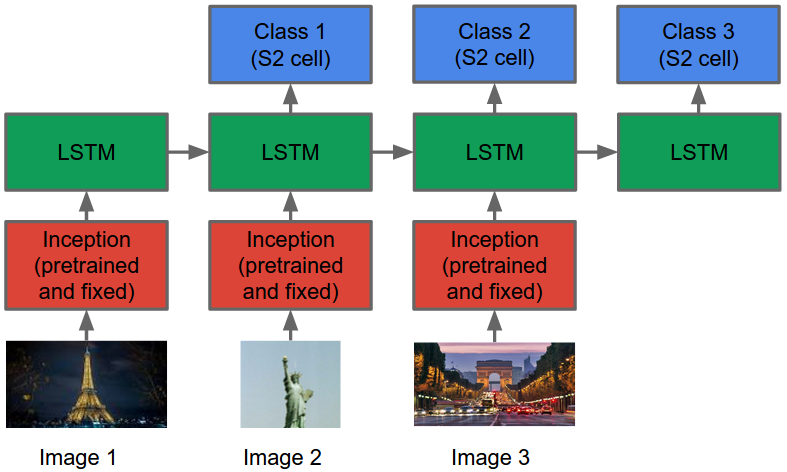}}
  \hspace{8pt}
\subfloat[\label{fig:lstm_repeated}]{\includegraphics[width=.32\linewidth]{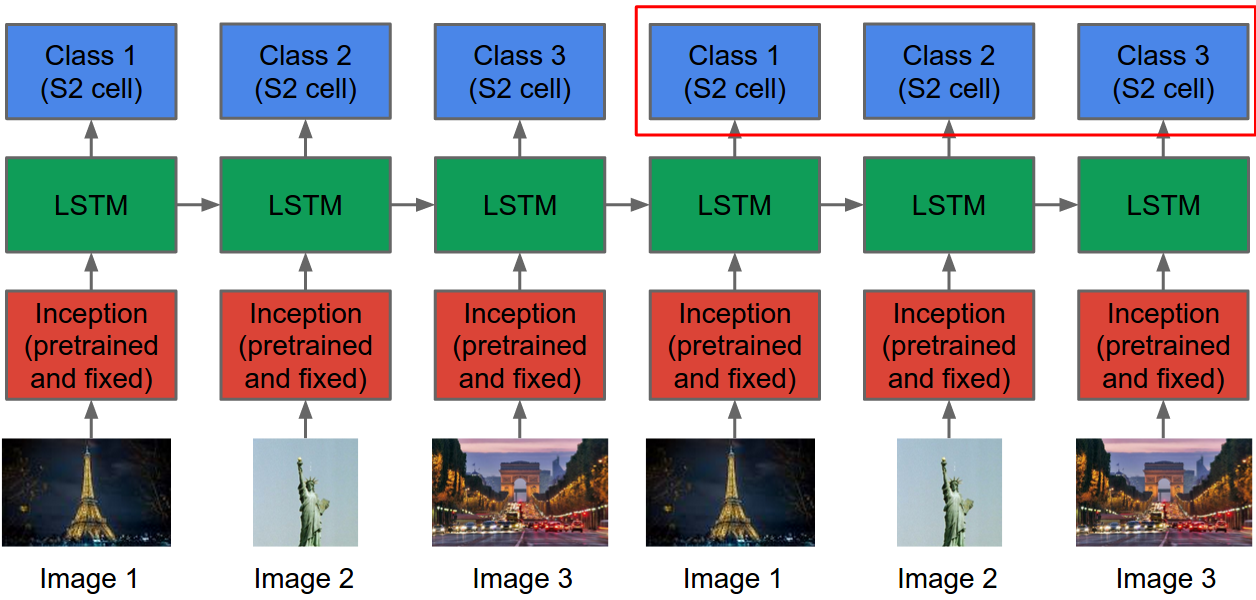}}
\hspace{8pt}
 %.48
 \subfloat[\label{fig:blstm_model}]{\includegraphics[width=.16\linewidth]{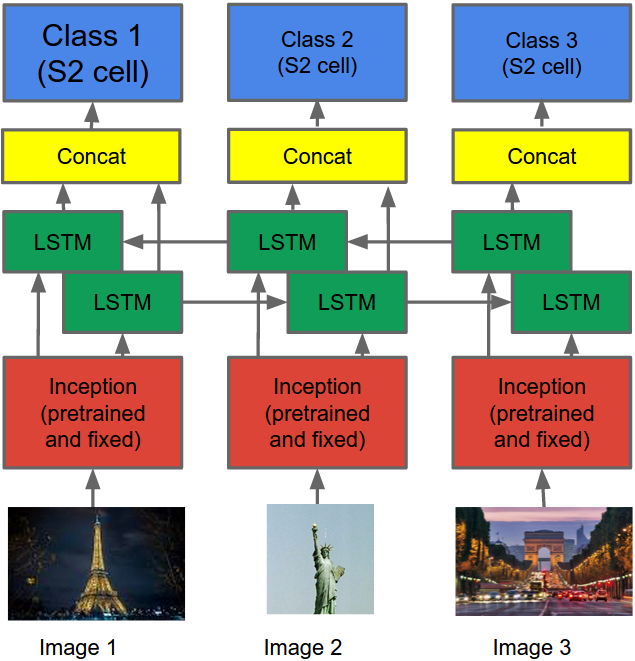}}
 \caption{Time-unrolled diagrams of the PlaNet LSTM models. (a)
   Basic model. (b) Label offset. (c) Repeated sequence. The first
   pass is used to generate the state inside the LSTM, so we only use
   the predictions of the second pass (red box). (d) Bi-directional
   LSTM.}
\end{figure*}
The basic structure of our model is as follows
(Fig.~\ref{fig:lstm_model1}): Given an image, we extract an embedding
vector from the final layer before the SoftMax layer in PlaNet.
This vector is fed into the LSTM unit. The output vector of the LSTM
is then fed into a SoftMax layer that performs the classification into
S2 cells. We feed the images of an album into the model in
chronological order. For the Inception part, we re-use the
parameters of the single-image model. During training, we keep the
Inception part fixed and only train the LSTM units and the SoftMax
layer.

\begin{table}[t]
  \centering
    \footnotesize
  \setlength\tabcolsep{3pt}
  \begin{tabular}{lrrrrr}
    \hline
    \textbf{Method} & \textbf{\shortstack{Street\\1 km}} & \textbf{\shortstack{City\\25 km}} & \textbf{\shortstack{Region\\200 km}} & \textbf{\shortstack{Country\\750 km}} & \textbf{\shortstack{Continent\\2500 km}} \\
    \hline
    PlaNet & 14.9\% & 20.3\% & 27.4\% & 42.0\% & 61.8\% \\
    PlaNet avg & 22.2\% & 35.6\% & 51.4\% & 68.6\% & 82.7\% \\
    LSTM & 32.0\% & 42.1\% & 57.9\% & 75.5\% & 87.9\% \\
    \hline
    LSTM off1 & 30.9\% & 41.0\% & 56.9\% & 74.5\% & 85.4\% \\
    LSTM off2 & 29.9\% & 40.0\% & 55.8\% & 73.4\% & 85.9\% \\
    LSTM rep & 34.5\% & 45.6\% & 62.6\% & 79.3\% & 90.5\% \\
    \hline
    LSTM rep 25 & 28.3\% & 37.5\% & 49.9\% & 68.9\% & 82.0\% \\
    BLSTM 25 & 33.0\% & 43.0\% & 56.7\% & 73.2\% & 86.1\% \\
    \hline
  \end{tabular}
  \caption{Results of PlaNet LSTM on Google+ photo albums.
Percentages are the fraction of images in the dataset localized within
the respective distance.}
  \label{tab:basic-lstm-results}
\end{table}

\PAR{Results.}
We compare this model to the single-image PlaNet model and a
baseline that simply averages the single-image PlaNet predictions of all
images in an album and assigns the average to all images. The results
are shown in Tab.~\ref{tab:basic-lstm-results} (first 3 rows).
Averaging within albums ('PlaNet avg') already yields a
significant improvement over single-image PlaNet (45.7\%
relative on street level), since it transfers more confident
predictions to ambiguous images. However, the LSTM model clearly
outperforms the averaging technique (50.5\% relative improvement on
the street level). Visual inspection of results showed that if an
image with high location confidence is followed by several images with
lower location confidence, the LSTM model assigns the low-confidence
images locations close to the high-confidence image. Thus, while the
original PlaNet model tends to ``jump around'', the LSTM model
tends to predict close-by locations unless there is strong evidence of
a location change. The LSTM model outperforms the averaging baseline
because the baseline assigns all images in an album the same
confidences and can thus not produce accurate predictions for albums
that include different locations (such as albums of trips).

A problem with this simple LSTM model is that many albums contain a
number of images in the beginning that contain no helpful visual
information. Due to its unidirectional nature, this model cannot fix
wrong predictions that occur in the beginning of the sequence after
observing a photo with a confident location. For this reason, we now
evaluate a model where the LSTM ingests multiple photos from the album
before making its first prediction.

\PAR{Label offset.}
The idea of this model is to shift the labels such that inference is postponed for several time steps (Fig.~\ref{fig:lstm_offset}). The main motivation under this idea is that this model can accumulate information from several images in a sequence before making predictions. Nevertheless, we found that using offsets does not improve localization accuracy (Tab.~\ref{tab:basic-lstm-results}, LSTM off1, LSTM off2). We assume this is because the mapping from input image to output labels becomes more complex, making prediction more difficult for all photos, while improving predictions just for a limited amount of photos. Moreover, this approach does not solve the problem universally: For example, if we offset the label by 2 steps, but the first image with high location confidence occurs only after 3 steps, the prediction for the first image will likely still be wrong. To fix this, we now consider models that condition their predictions on all images in the sequence instead of only previous ones.

\PAR{Repeated sequences.}
We first evaluate a model that was trained on sequences that had been constructed by concatenating two instances of the same sequence (Fig.~\ref{fig:lstm_repeated}). For this model, we take predictions only for the images from the second half of the sequence (\ie the repeated part). Thus, all predictions are conditioned on observations from all images. At inference time, passing the sequence to the model for the first time can be viewed as an \emph{encoding} stage where the LSTM builds up an internal state based on the images. The second pass is the \emph{decoding} stage where at each image, the LSTM makes a prediction based on its state and the current image. Results show that this approach outperforms the single-pass LSTMs (Tab.~\ref{tab:basic-lstm-results}, LSTM rep), achieving a 7.8\% relative improvement at street level, at the cost of a twofold increase in inference time. However, by visually inspecting the results we observed a problem with this approach: if there are low-confidence images at the beginning of the sequence, they tend to get assigned to the last confident location in the sequence, because the model learns to rely on its previous prediction. Therefore, predictions from the end of the sequence get carried over to the beginning.

\PAR{Bi-directional LSTM.}
A well-known neural network architecture that conditions the predictions on the whole sequence are bi-directional LSTM (BLSTM) \cite{Graves05NN}. This model can be seen as a concatenation of two LSTM models, where the first one does a forward pass, while the second does a backward pass on a sequence (Fig.~\ref{fig:blstm_model}). Bi-directional LSTMs cannot be trained with truncated back-propagation through time \cite{Elman90CS} and thus require to unroll the LSTMs to the full length of the sequence. To reduce the computational cost of training, we had to limit the length of the sequences to 25 images. This causes a decrease in total accuracy since longer albums typically yield higher accuracy than shorter ones. Since our experiments on this data are not directly comparable to the previous ones, we also evaluate the repeated LSTM model on sequences truncated to 25 images. As the results show (Tab.~\ref{tab:basic-lstm-results}: LSTM rep 25, BLSTM 25), BLSTMs clearly outperform repeated LSTMs (16.6\% relative improvement on street level). However, because they are not tractable for long sequences, the repeated model might still be preferable in practice.

\section{Conclusion}
We presented PlaNet, a CNN for image geolocation. Regarding problem as classification, PlaNet produces a probability distribution over the globe. This allows it to express its uncertainty about the location of a photo and assign probability mass to potential locations. While previous work mainly focused on photos taken inside cities, PlaNet is able to localize landscapes, locally typical objects, and even plants and animals. Our experiments show that PlaNet far outperforms other methods for geolocation of generic photos and even reaches superhuman performance.
We further extended PlaNet to photo album geolocation by combining it with LSTMs. Our experiments show that using contextual information for image-based localization makes it reach 50\% higher performance than the single-image model.

{\small
\bibliographystyle{ieee}
\bibliography{abbrev_short,planet}
}

\end{document}